%% file: cvpr_main.tex
\documentclass[10pt,twocolumn,letterpaper]{article}

\usepackage[pagenumbers]{cvpr} 

\input{header}

\usepackage[accsupp]{axessibility}  
\usepackage[pagebackref,breaklinks,colorlinks]{hyperref}

\usepackage[capitalize]{cleveref}
\crefname{section}{Sec.}{Secs.}
\Crefname{section}{Section}{Sections}
\Crefname{table}{Table}{Tables}
\crefname{table}{Tab.}{Tabs.}

\begin{document}

\title{PartGlot: Learning Shape Part Segmentation from Language Reference Games}

\author{Juil Koo\textsuperscript{1} $\quad$
Ian Huang\textsuperscript{2} $\quad$
Panos Achlioptas\textsuperscript{2,3} $\quad$
Leonidas Guibas\textsuperscript{2} $\quad$
Minhyuk Sung\textsuperscript{1} \\
\textsuperscript{1}KAIST $\quad$ \textsuperscript{2}Stanford University $\quad$ \textsuperscript{3}Snap Inc.}

\twocolumn[{%
\renewcommand\twocolumn[1][]{#1}%
\maketitle
\vspace{-0.4cm}
\begin{center}
\centering
\captionsetup{type=figure}
\includegraphics[width=\textwidth]{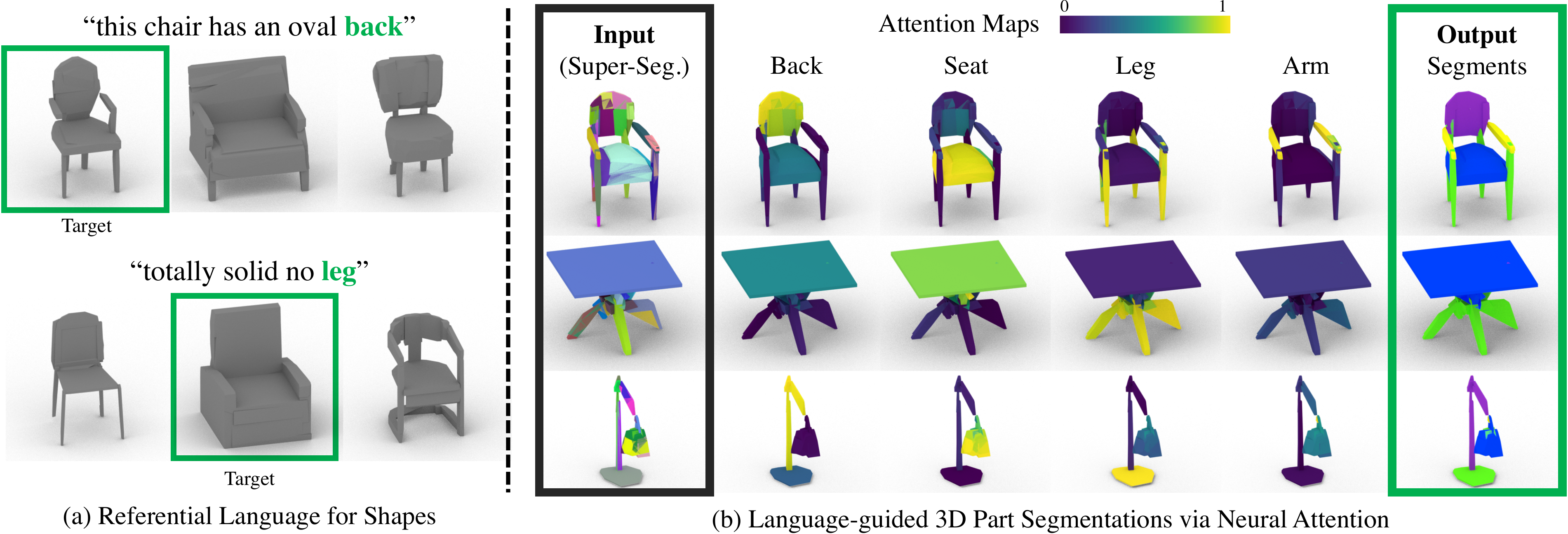}
\caption{\textbf{Overview.} On the left, we present examples of \textit{referential language} distinguishing the \textbf{shape} of a \emph{target} geometry (enclosed inside a green box) from two \emph{distractor} objects.
Using such language, our proposed task is to directly predict semantic part segments of 3D objects.
On the right, we present the examples of our segmentation results:
given unsupervised 3D super-segments of a shapes and the referential language, we learn a set of \textbf{attention maps} that corresponds to semantic shape parts, discovered \textbf{solely} by solving the language-reference problems of identifying the target shape. Tapping on the zero-shot learning capacity of natural language learners, and the shared part composition of common objects, we find examples of \textbf{zero-shot} segmentations on a table and lamp objects, extracted from learners and language concerning \textit{only} chair-based comparisons (the second and third rows).}
\label{fig:teaser}
\end{center}
}]

\input {sections/00_Abstract}
\input {sections/01_Introduction}

\input {sections/02_Related_Work}
\input {sections/03_Method}

\input {sections/04_Experiment_Result}
\input {sections/05_Conclusion}

\vspace{-0.5\baselineskip}
\paragraph{Acknowledgments}
This work was supported in part by NRF grant (2021R1F1A1045604) and NST grant (CRC 21011) funded by the Korea government(MSIT), Technology Innovation Program (20016615) funded by the Korea government(MOTIE), and grants from the Adobe and KT corporations. The Stanford team also acknowledges the support of ARL grant W911NF2120104, Vannevar Bush and TUM/IAS faculty fellowships, and grants from the Adobe and Snap Corporations.

{\small
\bibliographystyle{ieee_fullname}
\bibliography{egbib}
}

\renewcommand{\thesection}{A}
\renewcommand{\thetable}{A\arabic{table}}
\renewcommand{\thefigure}{A\arabic{figure}}

\newif\ifpaper
\papertrue

\clearpage
\newpage

\section*{Appendix}
\input {sections/Supplementary}

\end{document}

%% file: header.tex
\usepackage{amsmath}
\usepackage{amssymb}
\usepackage{booktabs}
\usepackage{dsfont}
\usepackage{graphicx}
\usepackage{makecell}
\usepackage{multirow}
\usepackage{pifont}
\usepackage[group-separator={,}]{siunitx}
\usepackage{tabularx}
\usepackage{xcolor}

\usepackage{catchfile}
\usepackage{longtable}

\newcolumntype{Y}{>{\centering\arraybackslash}X}

\DeclareMathOperator*{\argmax}{arg\,max}

\definecolor{color_1}{RGB}{255,0,128}
\definecolor{color_2}{RGB}{0,128,128}
\definecolor{color_3}{RGB}{0,128,0}
\definecolor{color_4}{RGB}{128,0,0}
\definecolor{color_5}{RGB}{128,0,128}

%% file: sections/00_Abstract.tex
\begin{abstract}
\vspace{-\baselineskip}
We introduce PartGlot, a neural framework and associated architectures for learning semantic part segmentation of 3D shape geometry, based solely on part referential language. We exploit the fact that linguistic descriptions of a shape can provide priors on the shape's parts -- as natural language has evolved to reflect human perception of the compositional structure of objects, essential to their recognition and use.
For training we use ShapeGlot's paired geometry / language data collected via a reference game where a speaker produces an utterance to differentiate a target shape from two distractors and the listener has to find the target based on this utterance~\cite{Achlioptas:2019}. Our network is designed to solve this target multi-modal recognition problem, by carefully incorporating a Transformer-based attention module so that the output attention can precisely highlight the semantic part or parts described in the language. Remarkably, the network operates without any direct supervision on the 3D geometry itself. Furthermore, we also demonstrate that the learned part information is generalizable to shape classes unseen during training.
Our approach opens the possibility of learning 3D shape parts from language alone, without the need for large-scale part geometry annotations, thus facilitating annotation acquisition. The code is available at \href{https://github.com/63days/PartGlot/}{https://github.com/63days/PartGlot}.

\end{abstract}

%% file: sections/01_Introduction.tex
\section{Introduction}
\label{sec:introduction}


Object perception is often based on structural abstractions --- the decomposition of an object into its parts and their inter-relationships~\cite{deformable_part_models,dubrinova_parts, part-whole-relations}. Natural language reflects this aspect of human perception of 3D shapes -- when a human is asked to describe an object, the description usually involves words \emph{naming} parts and expressions about part attributes 
and their relationships.
This implies that, conversely, language descriptions of an object can provide priors on the compositional structure of the object geometry, including the identity of its components or parts.
In this paper, we study the interplay between these two very different modalities, geometry and language, and how it can guide learning shape structure and parts.

ShapeGlot~\cite{Achlioptas:2019} explored the interplay between natural language and object geometry for the task of differentiating objects. It proposed a way to design a crowd-sourcing task to elicit more part-related referential language (utterances) about objects from users, based on a \emph{reference game}. Specifically, one user (the speaker) is shown 
three related objects
(a ``target" shape and two ``distractor" shapes) and is asked to describe
how the target is \textit{different} from the distractors. A second user (the listener) is then
asked to select the one described by the first user.
An interesting aspect of this work is that even though in training the (referential) neural networks are only given holistic
shape representations
with no part information whatsoever, they learn to depend heavily on part-related words and the corresponding visual parts of objects. 


Motivated by this initial observation and using the same data, our work investigates how well a neural network can connect part names in the utterances to specific regions in the geometry of the 3D shapes. We show the remarkable fact that \textbf{geometric object part structure can emerge from language alone,
without any direct geometric supervision on part segments},
highlighting the deep ties between language and geometry.
In other words, we can discover semantic part segments on the geometry by exploiting solely referential language data.
Even the language data we use is \emph{pragmatic}, not guided by any comprehensive partonomy as done by previous work~\cite{vlgrammar}, but merely focusing on describing shape differences.

Our framework is based on a variant of the neural listener pipeline in ShapeGlot, taking a language utterance plus three 3D shapes in point cloud format and predicting the probability of how likely each of the shapes is to be the target described by the utterance. For this learning task, we explore the application of a Transformer-based attention module~\cite{Vaswani:2017} to learn the region corresponding to each part described in the utterance as attention focus. Simply plugging in an attention module, however, does not produce any meaningful regions aligned with the semantic parts.
Hence, we make several important changes that lead the network to learn meaningful part segmentation masks as a byproduct of learning to identify the target shape.
Our experimental results demonstrate that the essential architectural components in our network significantly improve the performance of part segmentation.
Also, in the case when the full set of part names are given at training time, we show that this additional information can be leveraged to better detect and segment parts.
Furthermore, we show that our network can generalize to out-of-distribution categories of shapes -- specifically, with training done on \emph{Chairs}, good semantic masks can be extracted out of instances of \emph{Tables} and \emph{Lamps}.

Beyond studying the capability of neural networks to jointly understand language and shape, this work also suggests a new potential way to collect data for object part segmentation.  Object or scene  segmentation is a fundamental problem in many vision tasks, but the advance of learning-based segmentation techniques is gated by the availability of large-scale human segmentation annotations of 2D images or 3D models. Particularly for 3D, collecting manual annotations on 3D objects requires a huge amount of human effort and cost. In contrast to this, uttering a language description is a much more natural way for people to provide information about object structure and geometry. We hope to see a lot more work on how 3D segmentation can be improved using the language description of objects, without direct geometry supervision.

%% file: sections/02_Related_Work.tex
\section{Related Work}
\label{sec:related_work}

\paragraph{Language and Shape}

Works that explore the intersection between language and geometry have taken many forms, from resolving language references \cite{Achlioptas:2019, referit3d, language_grounding_with_3D_objects}, to generating language descriptions of a shape \cite{Achlioptas:2019,vlgrammar}, to generating a shape given a language description \cite{CLIPForge, static_and_animated_3D_scene_generation}. Most relevant to our work are the ones that attempt the language reference game, where the task is to select based on a language description a target shape out of a set of potential candidates either in a collection of individual 3D shapes \cite{Achlioptas:2019, language_grounding_with_3D_objects} or within a scene \cite{referit3d, language_refer, 3dvg_transformer, instance_refer,  text_guided_graph_neural_networks, towers_of_babel}. While most of these works treat the reference game as a classification problem on the set of candidates, \cite{text_guided_graph_neural_networks} outputs a segmentation mask over the scene. However, unlike our method, their work 1) applies to 3D scenes instead of individual shapes, and 2) requires full supervision for the segmentation task. To the best of our knowledge, our work is the first to derive part-level segmentation masks from spatial attention as a byproduct of learning to play the language reference game.


\paragraph{Transformers}


Not only have Transformers demonstrated superior performance in several tasks \cite{Hudson:2021, Vaswani:2017, Dosovitskiy:2021, Carion:2020, Xu:2015, Guo:2021}, but they also are characterized by interpretability of the attention map and can discover meaningful correspondences between different modalities \cite{Xu:2015}. In addition to being applied in the 2D visual domain \cite{Dosovitskiy:2021, Carion:2020, Guo:2021}, transformers have also been used in the 3D spatial domain for a variety of tasks, operating most commonly on point clouds. A variety of attention mechanisms has been introduced \cite{investigating_attention_mechanism_in_3D_point_cloud_object_detection}. For instance, \cite{pointr_diverse_point_cloud_completion_with_geometry_aware_transformers} adapts the transformer architecture for point cloud completion. Works like  \cite{point_attention_network_for_semantic_segmentation_of_3D_pointclouds, point_transformer, cloud_transformers} have shown superior performance on the semantic segmentation task by including modules that employ self-attention over point clouds. However, in all the above cases, the segmentation masks were developed with heavy supervision, and not extracted using the attention over the spatial domain. They furthermore do not attempt to leverage information from other modalities. Here, instead of using self-attention over only the spatial domain, we use cross-attention between multiple modalities --- a byproduct of learning language references --- for the segmentation task.


%

\paragraph{Self-Supervised or Weakly-Supervised Segmentation}

\cite{Zhu:2020} proposed a weakly-supervised shape co-segmentation method. Two key elements are the part prior network and low-rank loss. It first trains the part prior network to learn a part prior by denoising unlabeled segmented parts from random noise. From this pre-trained part prior network, the co-segmentation network is optimized to output consistent segmentations via low-rank loss. Low-rank loss regularizes the network to maximize the similarity of the part feature belonging to the same part by minimizing the rank of the matrix consisting of part features of the same part across all test shapes. \cite{Wang:2021} also utilizes two key elements introduced in \cite{Zhu:2020} for the fine-grained segmentation without part semantic tags. Those networks are trained in a label-agnostic manner, but still require segmentation information to train the part prior network. Our model does not require any part prior, but learns geometry from language on the fly.

\paragraph{Shape Decomposition}
Recently, there have been many works \cite{Chen:2020,tulsiani,Paschalidou:2019,Chen:2019,Paschalidou:2021,Deng:2020, spfn,Chen:2019,Sung:2020} for shape decomposition. \cite{tulsiani, Paschalidou:2019, Genova:2019} abstract a complex shape into multiple primitives, cuboids, superquadrics or Gaussians, by regressing the parameters of the primitive that fit to the target shape. \cite{Chen:2020, Deng:2020} decompose a shape as a collection of convexes. \cite{Chen:2019, Paschalidou:2021} learn an implicit field to represent the shape. Those works have demonstrated to abstract the shape into multiple primitives, but those primitives lack of semantics. So, they usually assigned the label for each primitive by hand in the test time. 

%% file: sections/03_Method.tex
\newcommand\Real{\mathbb{R}}
\newcommand\Feat{\mathbf{F}}

\section{Attention-Based Part Segmentation}
\label{sec:method}

\begin{figure*}
\includegraphics[width=\linewidth]{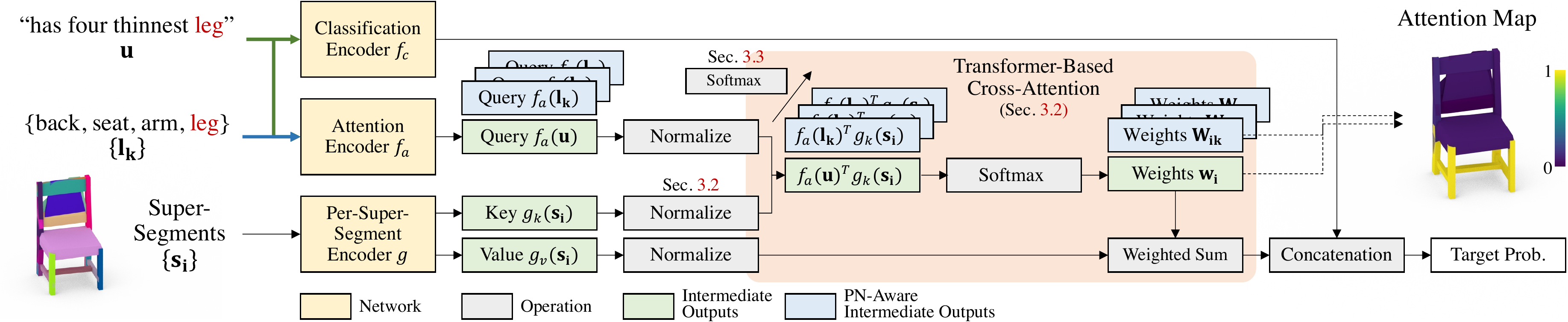}
\caption{A high-level overview of our architecture solving a reference problem. There are three main encoders: \emph{Classification} Encoder $f_c$, \emph{Attention} Encoder $f_a$ and Per-Super Segment Encoder $g$. The cross-attention module aggregates Per-Super-Segment features based on a query to output the shape feature. The concatenation of the output of the \emph{Classification} Encoder $f_c(\mathbf{u})$ and the shape feature is used to produce the final classification probability. The attention map contains the semantic part information corresponding to input language. At test time, we obtain part segments using the attention map of a template expression: : ``\texttt{a chair with \{part name\}}''.}
\label{fig:pipeline}
\end{figure*}

\subsection{Background and Overview} 
\label{sec:background}

We investigate the capability of a neural network (specifically, an attention module) to learn semantic parts of a 3D object solely from referential language about it without any explicit supervision of its part segmentations. To achieve this, we deploy a listening comprehension task similar to ShapeGlot's~\cite{Achlioptas:2019} as our basic learning objective. Specifically, given a triplet of shapes and an utterance differentiating one of them,
our main task is to learn to identify the referred target shape (Figure~\ref{fig:pipeline}).
For this task, a variety of viable neural network architectures can be designed to assign a probability to each shape indicating its congruence with the underlying utterance. This work is the first to demonstrate that by carefully incorporating an attention module over the 3D spatial domain of the visual stimulus (e.g., attention over unordered sets of 3D point clouds), our network can not only learn to identify the target shape but also discover 3D regions of the parts described in the utterance as a byproduct of solving the reference task.


We adapt the original neural network architecture in ShapeGlot to better facilitate our goal of recognizing and segmenting object parts with language alone. First, we focus on the application of neural listeners operating \textit{solely} on 3D geometric representations  -- 3D point clouds -- and ignore 2D image-based projections used in ShapeGlot.
Second, we also explore the effect of partitioning the input point cloud into subgroups, namely \emph{super-segments} (analogous to superpixels~\cite{Achanta:2012} in 2D), and cast the semantic part-prediction problem over those larger entities. Crucially, super-segments (groups of points) can be derived with a self-supervised approach (in our experiments, we use the output of BSP-Net~\cite{Chen:2020}); hence their utilization does not undermine our goal of annotation-free 3D part segmentations.
Third, we add a Transformer-based \cite{Vaswani:2017} attention module taking the utterance or a part name as a query. We also change the architecture of the geometry encoder to make the neural network seek an appropriate local region for attention; more details are described below. We investigate two different setups of the problem: with and without knowledge of the full set of part names during training.

\subsection{Part-Name-Agnostic (PN-Agnostic) Learning} 
\label{sec:pn_free}
We first describe a learning scenario where the set of part names is \emph{not} given during training. The association between part names (words) and regions in the 3D shape in this case has to be learned solely from the
connection between utterances (a single or multiple sentences) and the entire 3D shapes. 

The network architecture is illustrated in Figure~\ref{fig:pipeline}. The input utterance $\mathbf{u}$ is encoded into two encoders: 
\emph{attention} encoder $f_a(\cdot)$ which decides ``where to look'' and \emph{classification} encoder $f_c(\cdot)$ which determines ``whether it is the target shape or not''. For both encoders,
we use an utterance encoder similar to the one used in the ShapeGlot; 
the token codes of the words in a sentence are randomly initialized and then processed via an LSTM sequentially with the standard bilinear word attention mechanism \cite{bilinear_attention_mechanism}. The output of the \emph{attention} encoder $f_a(\mathbf{u})$
becomes the \emph{query} vector in the subsequent Transformer~\cite{Vaswani:2017}, and the output of the \emph{classification} encoder $f_c(\mathbf{u})$ is concatenated with the output of the Transformer (a weighted sum of the super-segment features) and is used to predict the classification probabilities.

For the three input shapes $\{o_1, o_2, o_3\}$, the target and two distractors, each of which is represented a set of super-segments $o=\{\mathbf{s_i}\}$, we extract a key $g_k(\mathbf{s_i})$ and a value $g_v(\mathbf{s_i})$ vector of each super-segment $s_i$ using PointNet \cite{pointnet}. In the following single cross-attention layer,
the attention from the utterance $\mathbf{u}$ to each super-segment $\mathbf{s_i}$ is calculated by taking a dot product of the query and key --- let $\mathbf{x}$ be a vector where $\mathbf{x}_i=f_a(\mathbf{u})^T g_k(\mathbf{s_i})$ --- and then normalizing them over super-segments using a softmax:

\begin{align}
\vspace{-0.25\baselineskip}
    \mathbf{w_i} = \sigma(\mathbf{x})_i = \frac{e^{\mathbf{x}_i}}{\sum_i e^{\mathbf{x}_i}}.
    \label{eq:transformer}
\vspace{-0.25\baselineskip}    
\end{align}

The resulting probability distribution over the super-segments $\{ \mathbf{w_i}\}$ becomes the attention expected to indicate the part described in the utterance $\mathbf{\mathbf{u}}$. The value vectors $\{ g_v(\mathbf{s_i}) \}$ are then aggregated by taking the probabilities $\{ \mathbf{w_i}\}$ as weights in a weighted mean, concatenated with the output of the classification encoder $f_c(\mathbf{u})$, and fed to an MLP to predict the classification score of each object.

A crucial detail in this architecture is to \emph{normalize} the query $f_a(\mathbf{u})$, key $g_k(\mathbf{s_i})$, and value $g_v(\mathbf{s_i})$ vectors to have a unit norm. Although missing this normalization does not affect the accuracy of target shape discrimination, it largely influences the attention and helps align the attention to a semantic part in practice since the weights in the attention can vary according to the different norms of the value vectors $g_v(\mathbf{s_i})$. The effect of the normalization is shown in Section~\ref{sec:results}. Note that, in Equation~\ref{eq:transformer}, we also do not divide the dot product of the query and key by the square root of the vector dimension as typically done in Transformer since all the vectors are normalized.

One more important observation is that the method used to process \emph{set} data is critical. Following PointNet \cite{pointnet}, many neural networks processing set data use the idea of combining local features with a \emph{global} feature, which is created by aggregating all the local features using a symmetric function such as max-pool. In our pipeline, we find that the concatenation of the global feature to each super-segment feature results in totally meaningless attention since the Transformer does not need to attend to a specific region, as each point can provide global shape information to complete the reference task. Hence, all super-segments are processed independently with the shared network module.

At test time, we obtain the attention of a part by leveraging a similar setup as CLIP \cite{Radford:2021}; a template language expression is used as the input utterance. In our experiments, we specifically use an expression: ``\texttt{a chair with \{part name\}}''. Given a set of part names, a segment per part is achieved by taking super-segments whose probability of the part attention is higher than the probabilities of any other attentions.

\subsection{Part-Name-Aware (PN-Aware) Learning} 
\label{sec:pn_aware}
In the case when the set of part names $\{ \mathbf{l_k} \}$ is predefined at the training time, we leverage this additional supervision to better align the output attentions to the given set of parts. Note that there is still no part segmentation supervision, since only the set of part names is given. In this setup, we also assume that each utterance describes one and only one part in the given set.

From the architecture introduced in Section~\ref{sec:pn_free}, we first change \emph{attention} encoder $f_a(\cdot)$ to take not the input utterance $\mathbf{\mathbf{u}}$ but the part name $\mathbf{l}$ described in the utterance as the input. Hence, a single-layer MLP for the part name latent token is used instead of an LSTM. In the test time, we also do not need to use a template expression; the part name can be directly fed to the attention encoder. Also, since now the set of part names is given, we propose to \emph{jointly} normalize the attentions of different part names, which is essential
to improving attention-based part segmentation. We specifically collect the dot products of the query and key vectors $f_a(\mathbf{l_k})^T g_k(\mathbf{s_i})$ for all the part names $\{ \mathbf{l_k} \}$ and super-segments $\{ \mathbf{s_i} \}$. Let $\mathbf{X}$ be a matrix where $\mathbf{X_{ik}}=f_a(\mathbf{l_k})^T g_k(\mathbf{s_i})$.
Then, we apply softmax to $\mathbf{X}$ \emph{twice}; along the set of part names first (along $k$) and then along the super-segments (along $i$):

\begin{align}
\vspace{-0.25\baselineskip}
    \mathbf{Y_{ik}} &= \sigma(\mathbf{X_{i,:}})_k \\
    \mathbf{W_{ik}} &= \sigma(\mathbf{Y_{:,k}})_i,
    \label{eq:double_softmax}
\vspace{-0.25\baselineskip}
\end{align}

where $\sigma(\cdot)$ is the softmax, and $\mathbf{W} = \{ \mathbf{W_{ik}} \}$ is the final weights. The first \emph{additional} softmax along with the part names ($k$) plays the role of making $\mathbf{X_{ik}}$ be \emph{spikier} for each super-segment, enforcing a super-segment to belong to \emph{only one} part name. This can thus avoid overlaps across the attention maps of different part names. We empirically find that still the final attention weights should be normalized over the super-segments to achieve the best performance. We show a comparison across different cases of applying softmax in our ablation study (Section~\ref{sec:results}).

\paragraph{Regularization Based on Cross Entropy}
To further enforce \emph{partitioning} of the output segments --- ensuring that a point is assigned to \emph{one and only one} part name --- we introduce a regularization loss based on cross entropy. Given $\mathbf{Y}$ (the output of the first softmax) and 
for each super-segment $\mathbf{s_i}$, we find the part name $\mathbf{l_k}$ that gives the highest probability $\mathbf{Y_{ik}}$ and compute the cross entropy loss by considering that the part as the ground truth label:

\begin{align}
\vspace{-0.25\baselineskip}
   \mathcal{L}_{\text{CE}} = \sum_{i} \sum_{k} -\mathds{1} \left( k = \argmax_{k'}(\mathbf{Y_{ik'}}) \right) \log(\mathbf{Y_{ik}})
   \label{eq:ce_loss}
\vspace{-0.25\baselineskip}
\end{align}

In addition to the double softmax, the regularization loss makes $\mathbf{Y}$ even spikier and further avoids overlaps across the attention maps of different part names. The ablation study in our experiments analyzes the effect in practice (Section~\ref{sec:results}).

%% file: sections/04_Experiment_Result.tex
\newcommand\cic{\textit{\textbf{C}i\textbf{C}}}
\newcommand{\cmark}{\ding{51}}%
\newcommand{\xmark}{\ding{55}}%

\section{Experiments}
\label{sec:experiment}

\subsection{Dataset and Evaluation }

In our experiments, we use the Chair in Context (\cic) dataset introduced in ShapeGlot~\cite{Achlioptas:2019}. \cic~includes sets of triplets of \emph{chairs} from ShapeNet~\cite{shapenet2015} (a target and two distractors) and an
utterance for the target chair, created by human speakers playing the grounded reference game.

\paragraph{Utterance Preprocessing}
We first preprocess the utterances of \cic~by fixing typos, converting plural nouns to singular nouns, and dividing a compound word into single words, e.g., ``armrest'' to ``arm rest''.
For the PN-Aware setup, we choose the following four part names as the given set: \emph{back}, \emph{seat}, \emph{leg}, and \emph{arm}, which are also chair part segments annotated in ShapeNet~\cite{shapenet2015}. We also only use chair triplets in \cic~ where their associated utterance describes only one of these parts.
After the preprocessing, the dataset contains 40,660 sets and 4,509 unique shapes.
We split the sets into train, validation, and test datasets with an 80\%-10\%-10\% ratio.
Since the numbers of the utterances describing each part are imbalanced, during training, we sample the utterances with the probabilities inversely proportional to the numbers of each part utterances.

\paragraph{Super-Segment Generation}
\begin{figure}[!h]
\begin{center}
\vspace{-\baselineskip}
\includegraphics[width=\columnwidth]{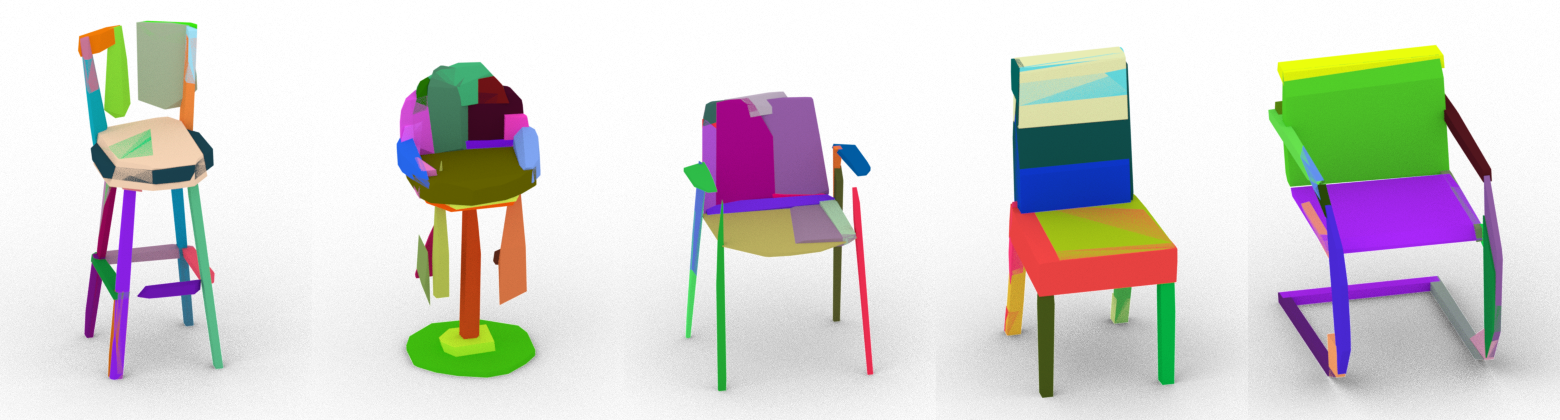}
\end{center}
\vspace{-\baselineskip}
\caption{Super-segments generated by a pretrained model of BSP-Net \cite{Chen:2020}. Distinct colors are randomly assigned to different super-segments.}
\label{fig:bsp-net}
\vspace{-\baselineskip}
\end{figure}
\input{tables/table_stats}
The super-segments of each shape are produced by a pre-trained BSP-Net \cite{Chen:2020}, provided by the authors; see examples in Figure~\ref{fig:bsp-net}. Then, each super-segment is represented with a small set of points, generated by randomly sampling 2,048 points over the entire shape and assigning them to super-segments based on proximity --- a point is assigned to \emph{one and only one} super-segment whose signed distance to the point is the minimum, and thus the super-segments \emph{partition} the point cloud.
See Table~\ref{tbl:data_stats} for the statistics of the number of super-segments and the number of points in each super-segment.
We further sample the points per super-segment so that the maximum number of points becomes \num{512}.


\paragraph{Segmentation Evaluation}
At test time, we obtain segments of the four parts --- \emph{back}, \emph{seat}, \emph{leg}, and \emph{arm} --- as the attention. Depending on whether the setup is PN-Aware or PN-Agnostic, either the template sentence mentioned in Section~\ref{sec:pn_free} or the part name itself is fed to the \emph{attention} encoder $f_a(\cdot)$ and used to generate the attention. The super-segments are assigned to the part name with the highest probability in the attention. The segmentation are evaluated based on ground truth part segmentation annotated in ShapeNet~\cite{shapenet2015}. The standard mIoU is used as the evaluation metric of the segmentation. The \emph{average} mIoU indicates taking mean \emph{per instance} and averaging over the shapes.

\subsection{Results}
\label{sec:results}

\input{tables/table_all_results}


\input{figures/figure_all_results}


The quantitative and qualitative results of our experiments are summarized in Table~\ref{tbl:results} and Figure~\ref{fig:results}.
We first show two comparisons PN-Agnostic (Section~\ref{sec:pn_free}) vs. PN-Aware (Section~\ref{sec:pn_aware}), and super-segments vs. points.
We then show the results of the ablation study for each crucial component in our pipeline. We also show the results of analyzing the effect of few-shot learning when assuming that the ground truth part segments are annotated in a few shapes.
In the end, we also demonstrate that our framework learns general part information that can be transferred to other shape classes (e.g., Tables and Lamps), and we also visualize the word attention in the utterance encoding.

\paragraph{PN-Agnostic vs. PN-Aware}
We first compare the two cases described in Section~\ref{sec:method}: when leveraging the set of part names in training (PN-Aware) or not (PN-Agnostic). The mIoUs are reported in rows 1 and 2 of Table~\ref{tbl:results}.
While PN-Agnostic (row 1) works well in most cases, it particularly shows a low mIoU for \emph{arm} compared to the case of learning with part names (40.6 vs 70.4). The arm is an optional part that may not exist in some shapes, and by definition of mIoU (used in PointNet~\cite{pointnet}), it becomes zero when there even exists a \emph{single} super-segment assigned to \emph{arm} while \emph{arm} does not exist. We observe that such failure cases happen when the full set of part names are not leveraged during training (see the examples in the \emph{second} and \emph{eighth} row in Figure~\ref{fig:results}), although these cases are greatly reduced after the part names are used (PN-Aware) with our essential components in the network. The reasons are further analyzed in the ablation study below. Note that the accuracy of target shape classification is almost the same for both cases. The cross-part mIoUs are reported in the supplementary.

For the rest of the experiments, we show the results for PN-Aware with various setups.


\paragraph{Super-Segments vs. Points}
We also demonstrate the advantage of using super-segments as input in our pipeline. We compare our case with two baselines: 1) using the raw point cloud as input (row 3 in Table~\ref{tbl:results}) and 2) using the point cloud but \emph{projecting} the prediction result to the super-segments in test time (row 4 in Table~\ref{tbl:results}). For the second, each point belonging to a super-segment votes for the part names, and the super-segment takes the part name of the majority.
Row 6 in Table~\ref{tbl:results} shows an upper bound, when the part names are assigned to super-segments based on the ground truth segmentation of the underlying point cloud.
When comparing our case using super-segments (row 2 and 5 in Table~\ref{tbl:results} --- these two are the same) with these two cases, the mIoUs are significantly improved, and even our results are close to the upper bound.
The low mIOUs of the second case (projecting point results to super-segments) show the value of these super-segments to be used in the training process, rather than just used in a post-processing step. See also the fifth column of Figure~\ref{fig:results} for poor qualitative results. 
The target shape classification accuracy is also a bit increased when super-segments are used instead of points. 

\paragraph{Ablation Study}
\label{sec:ablation}
We also demonstrate through an ablation study that the details in our network pipeline are crucial for the part segmentation performance. From rows 7 to 12 in Table~\ref{tbl:results}, we report the results of the following cases (in order): 1) when the query $f_a(u)$, key $g_k(s_i)$, and value $g_v(s_i)$ vectors are not normalized, 2) when the softmax $\sigma$ is applied across the super-segments only, 3) across the part names only, 4) across super-segments \textit{first} and \textit{then} part names (a reverse order), 5) when adding a global feature to the feature of each super-segment, and 6) when not using the cross-entropy-based regularization loss (Equation~\ref{eq:ce_loss}). 

From the results in the table and also in Figure~\ref{fig:results}, we can draw several conclusions. First, the normalization of query, key, and value vectors and also the softmax $\sigma$ along with the part names before the super-segments improve overall mIoUs and particularly help detect optional parts accurately. See the arms in the sixth and seventh columns of Figure~\ref{fig:results} compared to ours in the third column. Interestingly, with these, the attention is improved to be better aligned with the semantic parts while the accuracy of target shape classification is slightly decreased. Second, it is crucial to have the double softmax in the order of part names first and then super-segments.
When the order is switched (the ninth column), or applying softmax only across part names (the eighth column), the overall quality of segmentation becomes worse; see the red circles in the eighth and ninth columns of Figure~\ref{fig:results} for some failure examples.
Third, as discussed in Section~\ref{sec:method}, the attention is not aligned with the semantic parts \emph{at all} when a global feature is concatenated to a local feature of super-segments (the tenth column in Figure~\ref{fig:results}). The global feature is obtained by max-pooling the local features, following the idea of PointNet~\cite{pointnet}. This result is obvious since the network can access the global shape information from any super-segment without carefully \emph{attending} to a specific region. Last, the cross-entropy-based regularization improves mIoUs particularly for \emph{seat}, which has the smallest utterances in the training dataset (2,215 out of 32,600), and also increases the target shape classification accuracy.

\paragraph{Few-Shot Learning}
\label{sec:few_shot}
\input{figures/figure_k_shot}
We further investigate whether part segment annotations on a few shapes can improve the segmentation accuracy in a few-shot learning setup. Here, we only consider the PN-Aware case and also assume that a few shapes (1, 8, and 32) are given with ground truth part segmentation. Note that 1, 8, and 32 are very small numbers compared to 4,509 number of entire shapes in the training dataset. We test exploiting the additional supervision by learning per-point classification with a cross entropy loss and the given annotated shapes after each epoch of the target shape discrimination task learning the attention. The results in Table~\ref{tbl:results} (rows 13-15) show improvements of mIoUs with the few-shot learning. Figure~\ref{fig:few_shot} also illustrates an example that the segmentation boundaries are refined even with a single-shot.

\subsection{Out-of-Distribution Test}
\label{sec:out_of_distribution}
\begin{figure}[!h]
\begin{center}
\vspace{-0.5\baselineskip}
\includegraphics[width=\columnwidth]{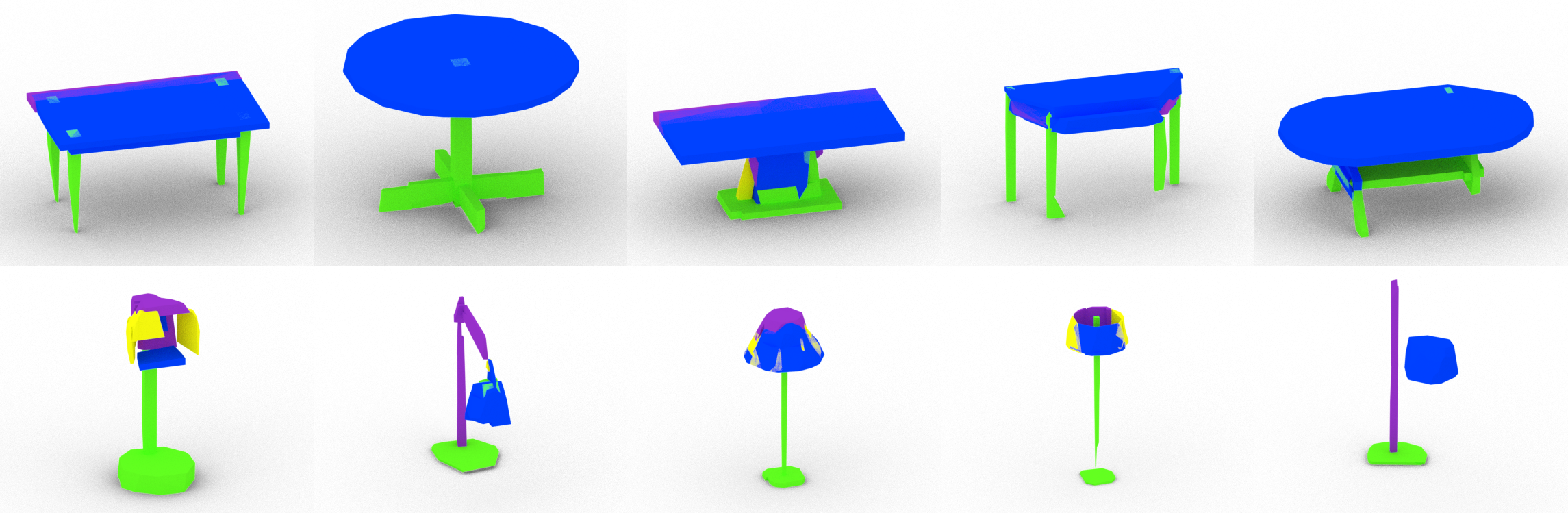}
\end{center}
\vspace{-\baselineskip}
\caption{Generalization to unseen shape categories. Each color indicates each predicted part as shown in Fig. \ref{fig:results}. In each category, the model predicts the part as the semantically matching part in a chair, predicting lower part as \emph{leg}.}
\label{fig:out_of_distribution}
\vspace{-\baselineskip}
\end{figure}
\input{tables/table_mIoU_all_pairs_others}

We experiment how much the part segment information learned from the Chairs in the \cic~dataset can be
\emph{zero-shot generalizable} to the other shape categories, namely, Tables and Lamps. Table~\ref{tbl:all_pairs_others} shows mIoUs across the parts of Chairs and the parts of Tables and Lamps. The results show very strong correlations between Chair \emph{seat} and Table \emph{top} as well as Chair \emph{leg} and Table \emph{leg}. Figure~\ref{fig:out_of_distribution} also clearly shows boundaries of the Table \emph{top} and \emph{leg} segments. Even the information about Chair parts is well-generalizable to Lamps, a category that is largely different geometrically from Chairs. The second part of the table illustrates that Lamp \emph{base} can be detected as Chair \emph{leg}, and also Lamp \emph{shade} is discriminated as Chair \emph{back} and \emph{seat}. The qualitative results are also shown in the second row of Figure~\ref{fig:out_of_distribution}.

\subsection{Word Attention Visualizations}
\label{sec:word_attention}
\begin{figure}[!h]
\vspace{-\baselineskip}
\begin{center}
\includegraphics[width=0.9\columnwidth]{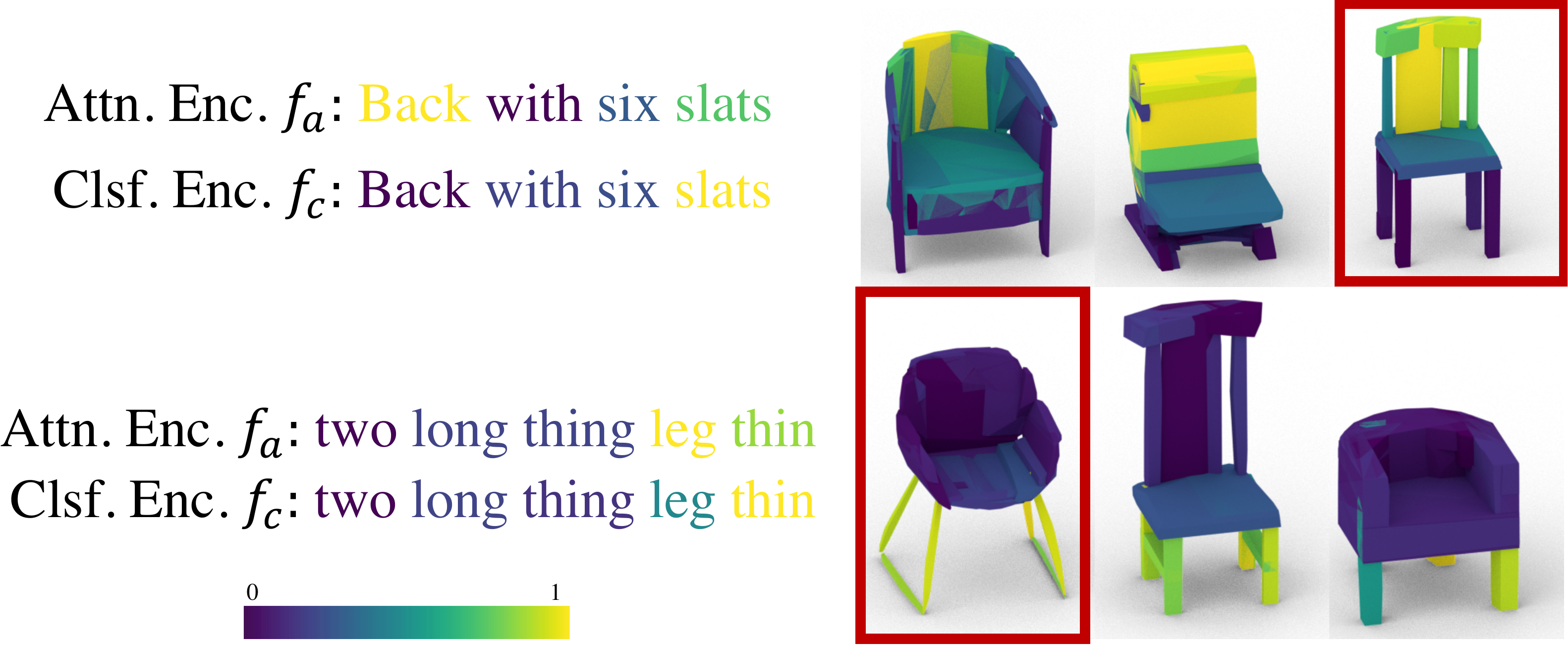}
\end{center}
\vspace{-\baselineskip}
\caption{Word Attention of PN-Agnostic. Two encoders attend different words in the utterance to play different roles: ``where to look" and ``what shape it should be".}
\label{fig:word_vis}
\vspace{-\baselineskip}
\end{figure}
Our utterance encoders, \emph{attention} encoder and \emph{classification} encoder (which use the same architecture as ShapeGlot \cite{Achlioptas:2019}) also learn attention over \emph{words}, and we visualize the word attention for some examples in Figure~\ref{fig:word_vis}. The color changes from dark blue to yellow when the attention weights for words increases from 0 to 1. Interestingly, the \emph{attention} encoder mainly attends the \emph{nouns} indicating the parts (the sentences above in each row), while the \emph{classification} encoder rather focuses on a general context (the sentences below).

%% file: tables/table_stats.tex
\begin{table}[h!]
\centering
\caption{Super-segment statistics.}
{\footnotesize
\begin{tabularx}{\linewidth}{>{\centering}m{2.4cm}|YYY}
\toprule
& Min. & Max. & Mean \\
\midrule
\# Super-Segs & 4 & 47.4 & 20.6 \\
\# Pts in Super-Segs & 0 & \num{1550.0} & 90.3 \\
\bottomrule
\end{tabularx}
\centering 
}
\label{tbl:data_stats}
\end{table}

%% file: tables/table_all_results.tex
\begin{table}[t!]
\centering
\caption{Quantitative Results of all experiments: [Id 1, 2] Comparison of two baselines; [Id 3-5]: Comparison of the input granularity; [Id 7-12]: ablation cases; [Id 13-15]: few-shot learning results. For each experiment, the model was selected with the highest classification accuracy on the validation set. \textbf{Bold} indicates the highest mIoU \emph{except} for the few-shot learning results.}
\footnotesize{
\setlength{\tabcolsep}{0.2em}
\renewcommand{\arraystretch}{1.0}
\begin{tabularx}{\linewidth}{>{\centering}m{0.3cm}|>{\centering}m{2.2cm}|Y|Y|Y|Y|Y|Y}
  \toprule
  \multirow{2}{*}{Id} &
  \multirow{2}{*}{Method} &
  \multicolumn{5}{c|}{Segmentation mIoU(\%)} &
  \multirow{2}{*}{\makecell{Classif.\\Acc.(\%)}}\\
  \cline{3-7}
  & & Back & Seat & Leg & Arm & Avg. & \\
  
  \midrule
  \multicolumn{8}{c}{PN-Agnostic (Sec.~\ref{sec:pn_free}) vs. PN-Aware (Sec.~\ref{sec:pn_aware})} \\
  \midrule
  1 & PN-Agnostic {\scriptsize(Ours)} & 82.2 & 78.8 & 75.5 & 40.6 & 69.3 & 61.6 \\
  2 & PN-Aware {\scriptsize(Ours)} & \textbf{84.9} & \textbf{83.6} & \textbf{78.9} & 70.4 & \textbf{79.4} & 61.5 \\
  
  \midrule
  \multicolumn{8}{c}{Points vs. Super-Segments (w/ PN-Aware)}\\
  \midrule
  3 & \footnotesize{Points} & 40.7 & 0.2 & 38.1 & 10.8 & 22.5 & 57.2  \\
  4 & \footnotesize{P $\rightarrow$ Sp.-Seg.} & 39.2 & 0 & 44.1 & 63.3 & 36.6 & 57.2 \\
  5 & \footnotesize{ \textbf{Sp.-Seg. {\scriptsize(Ours)}} } & \textbf{84.9} & \textbf{83.6} & \textbf{78.9} & 70.4 & \textbf{79.4} & 61.5 \\
  \midrule
  6 & \footnotesize{ Upper Bound* } & 89.8 & 88.9 & 85.2 & 92.3 & 89.1 & - \\
  
  \midrule
  \multicolumn{8}{c}{Ablation Study (w/ PN-Aware)} \\
  \midrule
  7 & w/o Unit Norm & 78.5 & 81.0 & 77.4 & 54.4 & 72.8 & 63.0 \\
  8 & $\sigma(\mathbf{X}) \rightarrow i$ & 80.8 & 77.5 & 75.3 & 56.6 & 72.5 & \textbf{63.4} \\
  9 & $\sigma(\mathbf{X}) \rightarrow k$ & 73.8 & 76.1 & 75.8 & \textbf{79.8} & 76.4 & 61.9 \\
  10 & $\sigma(\mathbf{X}) \rightarrow i \rightarrow k$ & 79.4 & 80.3 & 74.1 & 35.1 & 67.2 & 59.0 \\
  11 & w/ Global Feat. & 38.6 & 0.2 & 77.7 & 4.6 & 30.3 & 62.2 \\
  12 & w/o $\mathcal{L}_{\text{CE}}$ & 82.6 & 79.7 & 77.4 & 71.4 & 77.8 & 59.8 \\
  
  \midrule
  \multicolumn{8}{c}{Few-Shot Learning (w/ PN-Aware)} \\
  \midrule
  13 & k=1 & 85.5 & 83.5 & 78.4 & 73.2 & 80.1 &	59.4 \\
  14 & k=8 & 86.1 & 84.2 & 78.9 & 70.6 & 79.9 & 60.0 \\
  15 & k=32 &86.9 & 84.8 & 79.5 & 76.5 & 81.9 & 59.7 \\
  

  \bottomrule
\end{tabularx}
}
\vspace{-0.5\baselineskip}
\label{tbl:results}
\end{table}

%% file: figures/figure_all_results.tex
\setlength{\tabcolsep}{0em}
\def\arraystretch{0.0}
\begin{figure*}
{\scriptsize
\begin{tabularx}{\textwidth}{Y|YYYYYYYYYY}
GT &
\makecell{PN-Agnostic\\(Ours)} &
\makecell{PN-Aware\\(Ours)} &
Points &
P $\rightarrow$ Sp.-Seg. &
\makecell{w/o\\Unit Norm}&
$\sigma(\mathbf{X}) \rightarrow i$ &
$\sigma(\mathbf{X}) \rightarrow k$ &
\makecell{$\sigma(\mathbf{X})$\\$\rightarrow i \rightarrow k$} &
w/ Global Feat. &
w/o $\mathcal{L}_{\text{CE}}$ \\
  \midrule

\multicolumn{11}{c}{
\includegraphics[width=\textwidth]{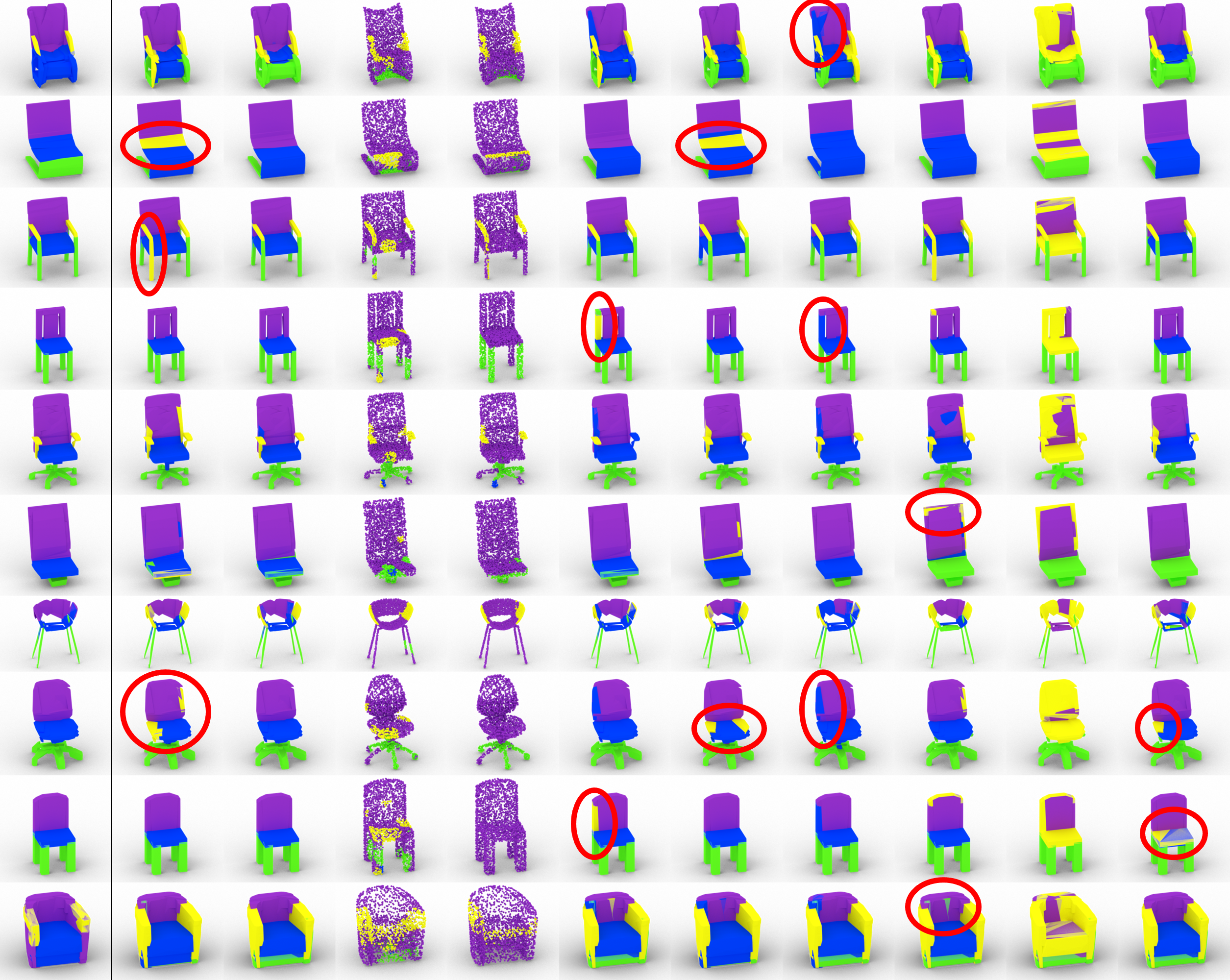}
}
\end{tabularx}
}
\caption{Qualitative examples of predicted part segmentations across variations of our method. Purple, blue, green and yellow indicate the prediction as \emph{back}, \emph{seat}, \emph{leg}, and \emph{arm}, respectively. The colors assigned to the super-segments in the ground truth column (GT) are computed based on ground truth point cloud part segmentation from ShapeNet \cite{shapenet2015}. Note that our PN-Agnostic and PN-Aware setups produced the best segmentation masks. Refer to the text for the details.}
\vspace{-0.5\baselineskip}
\label{fig:results}
\end{figure*}

%% file: figures/figure_k_shot.tex
\begin{figure}
{
\setlength{\tabcolsep}{0em}
\def\arraystretch{0.0}

{\scriptsize
\begin{tabularx}{\columnwidth}{Y|Y|YYY}
GT &
\makecell{Ours\\(PN-Aware)} &
k=1 &
k=8 &
k=32 \\
  \midrule
 \includegraphics[width=\linewidth]{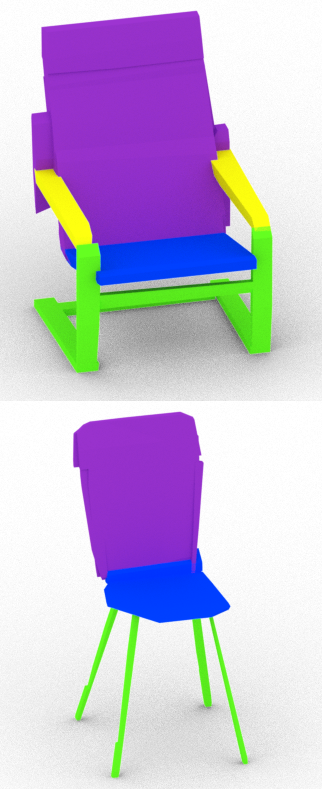} &
 \includegraphics[width=\linewidth]{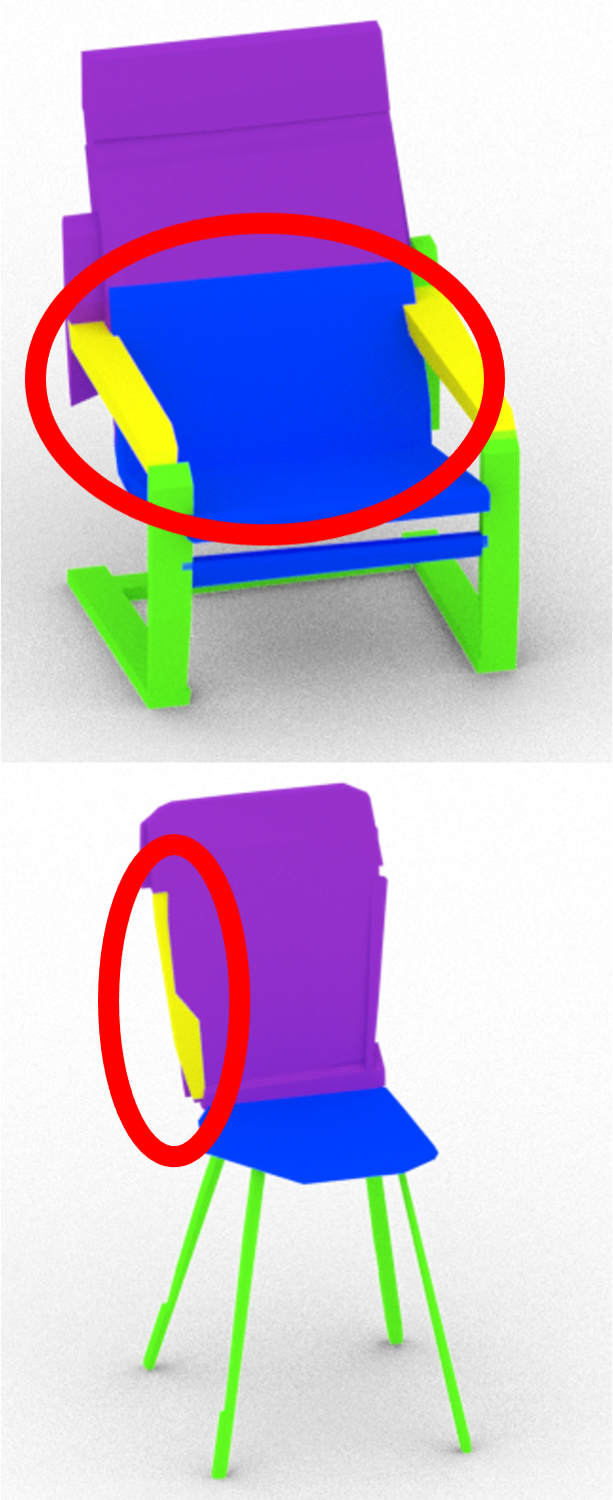} &
 \includegraphics[width=\linewidth]{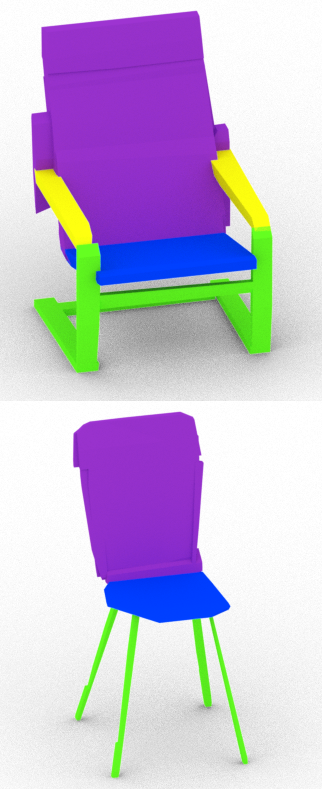} &
 \includegraphics[width=\linewidth]{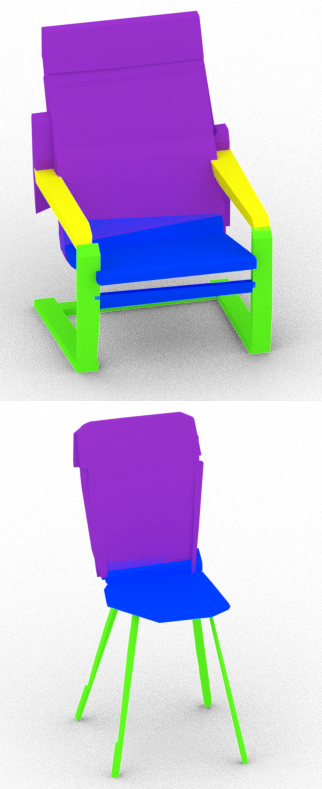} &
 \includegraphics[width=\linewidth]{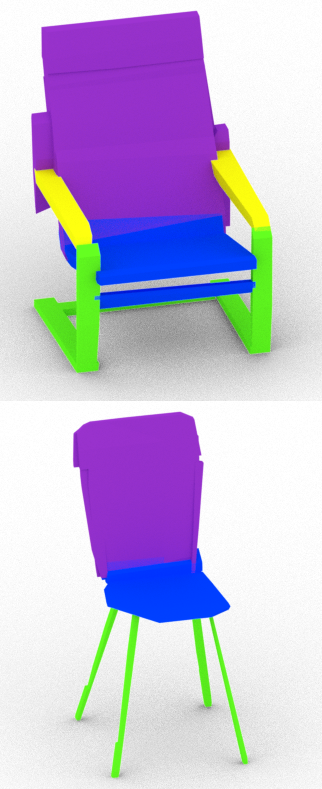}
\end{tabularx}
}
\caption{Qualitative improvements of using a few labelled examples. Even a very small amount of ground truth eliminates things that the model might be confused about without any supervision, such as the boundary between \emph{back} and \emph{seat}, and predicting the edge of the back as an arm.}
\label{fig:few_shot}
}
\end{figure}


%% file: tables/table_mIoU_all_pairs_others.tex
\begin{table}[t!]
\centering
\caption{Out-of-distribution quantitative results. Our model can also segment out-of-distribution shapes.
This table shows mIoUs with the attention maps learned with Chair parts and the part segments of the other classes in ShapeNet. Semantically corresponding parts have higher mIoUs, e.g., Chair \emph{leg} $\rightarrow$ Table \emph{leg} and Lamp \emph{base}.}

\footnotesize{
\setlength{\tabcolsep}{0.2em}
\renewcommand{\arraystretch}{1.0}
\begin{tabularx}{\linewidth}{Y|>{\centering}m{1.6cm}|Y|Y|Y|Y}
  \toprule
   \multicolumn{2}{c|}{\multirow{2}{*}{\makecell{Other\\Classes}}} & \multicolumn{4}{c}{Chair (w/ PN-Aware)} \\
   \cline{3-6}
   \multicolumn{2}{c|}{} & Back & Seat & Leg & Arm \\
  
   \midrule
   \multirow{3}{*}{Table} & Top & 11.0 & \textbf{78.2} & 1.2 & 3.5 \\
   & Leg & 4.5 & 2.8 & \textbf{66.2} & 11.0 \\
   & Connector & \textbf{26.5} & 3.2 & 2.1 & 15.7 \\
  
   \midrule
   \multirow{4}{*}{Lamps} & Base & 2.0 & 1.0 & \textbf{44.6} & 9.8 \\
   & Shade & 27.5 & \textbf{38.9} & 7.1 & 16.6 \\
   & Canopy & 4.9 & 7.0 & 5.1 & \textbf{20.8} \\ 
   & Tube & \textbf{21.4} & 7.7 & 20.6 & 2.2 \\
  
  \bottomrule
\end{tabularx}
}
\vspace{-0.5\baselineskip}
\label{tbl:all_pairs_others}
\end{table}

%% file: sections/05_Conclusion.tex
\section{Conclusion}
\label{sec:conclusion}
We proposed PartGlot, a framework learning part segmentation of 3D shape from linguistic descriptions. Without any direct supervision on part segmentation, our network classifying the target shape described by a given utterance can detect and segment part regions through an attention module. We not only introduced the first proposal of language-based 3D part segmentation but also designed a network curated for the emergence of part structure from the attention. We also proposed how predefined part names can be exploited in training to achieve the best performance. We finally demonstrated the part information learned by the network is transferable to other classes of shapes.




%% file: sections/Supplementary.tex
\ifpaper
  \newcommand{\refofpaper}[1]{\unskip}
  \newcommand{\refinpaper}[1]{\unskip}
  \newcommand{\suppSegDir}{supp_segmentations}
\else
  \makeatletter
  \newcommand{\manuallabel}[2]{\def\@currentlabel{#2}\label{#1}}
  \makeatother
  \manuallabel{sec:related_work}{2}
  \manuallabel{sec:pn_free}{3.2}
  \manuallabel{sec:pn_aware}{3.3}
  \manuallabel{sec:results}{4.2}
  \manuallabel{sec:out_of_distribution}{4.3}
  \manuallabel{tbl:results}{2}
  \manuallabel{fig:teaser}{1}
  \manuallabel{fig:results}{4}
  \newcommand{\refofpaper}[1]{of the main paper}
  \newcommand{\refinpaper}[1]{in the main paper}
  \newcommand{\suppSegDir}{supp_segmentations_resized}
\fi


\ifpaper
\else
In this supplementary material, we first further analyze the effect of using super-segments instead of points while varying the granularity of the super-segments (Section~\ref{sec:effect_super_segments}). Then, we demonstrate how much the attention module in our network affects the target shape discrimination in the reference games (Section~\ref{sec:effect_learning_attention}). We also show more results of the out-of-distribution test with Airplanes and Cars (quantitatively in Section~\ref{sec:ood_test_more} and qualitatively in Section~\ref{sec:more_segmentations}), and also analyze how much training data is needed to obtain meaning part segmentation results (Section~\ref{sec:effect_training_data_size}). The cross-part mIoUs are also reported in Section~\ref{sec:cross_part_mious}.
We also provide results when an additional regularization loss (group consistency loss introduced in AdaCoSeg~\cite{Zhu:2020}) is used (Section~\ref{sec:low_rank_reg}) and also a more recent text encoder (ALBERT~\cite{ALBERT}) is used instead of LSTM, while these variations do not change the results much.
We also experiment with finer-grained parts in PartNet~\cite{Mo:2019} and synthetic referential language and report the results in Section~\ref{sec:partnet_results}.
At the end, we provide implementation details (Section~\ref{sec:implementation_details}), and more qualitative results (Section~\ref{sec:more_segmentations}) and comparisons (Section~\ref{sec:more_comparisons}).
\fi

\subsection{Effect of Granularity of Super-Segments}
\label{sec:effect_super_segments}
\vspace{-0.5\baselineskip}
\begin{figure}[h]
\includegraphics[width=0.49\linewidth]{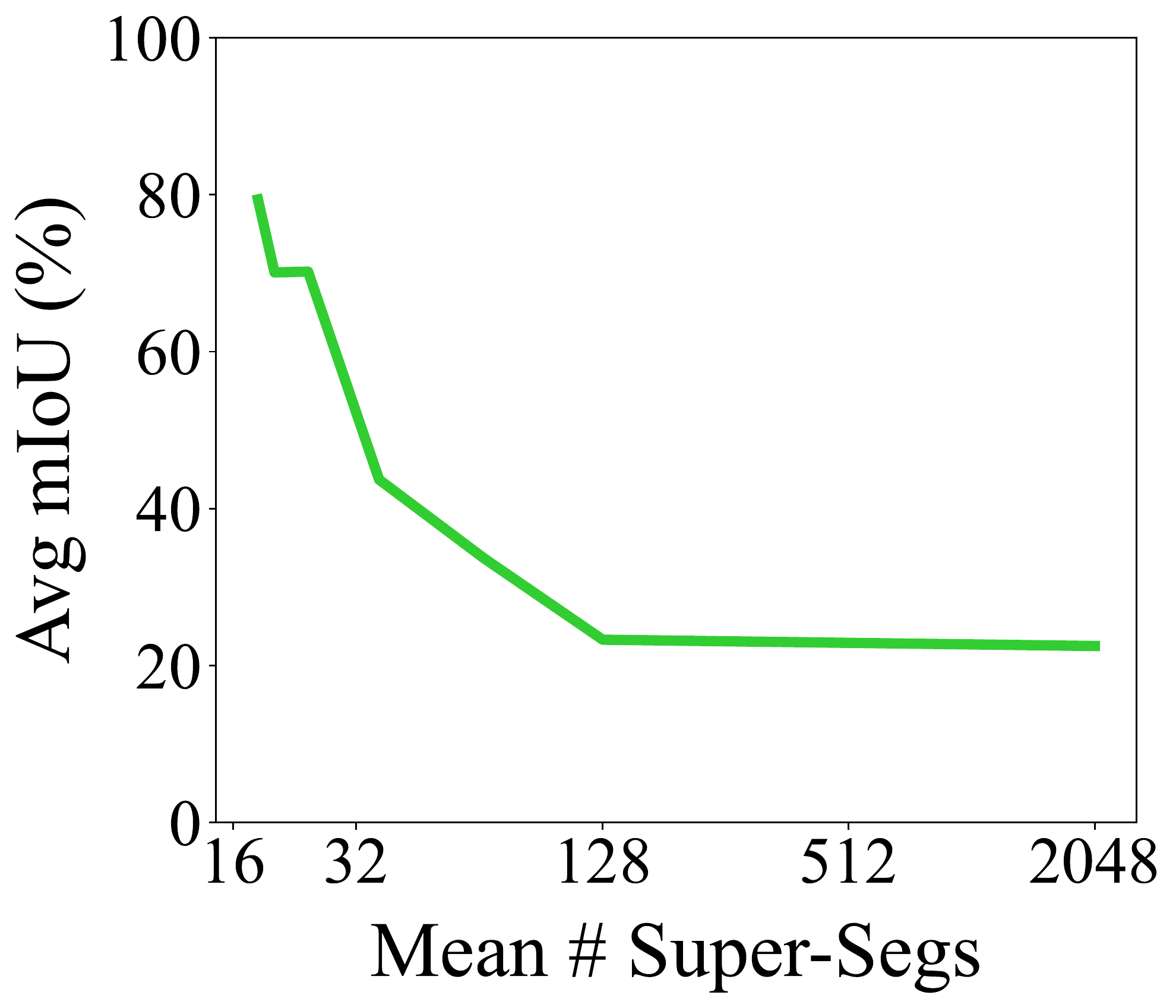}
\includegraphics[width=0.49\linewidth]{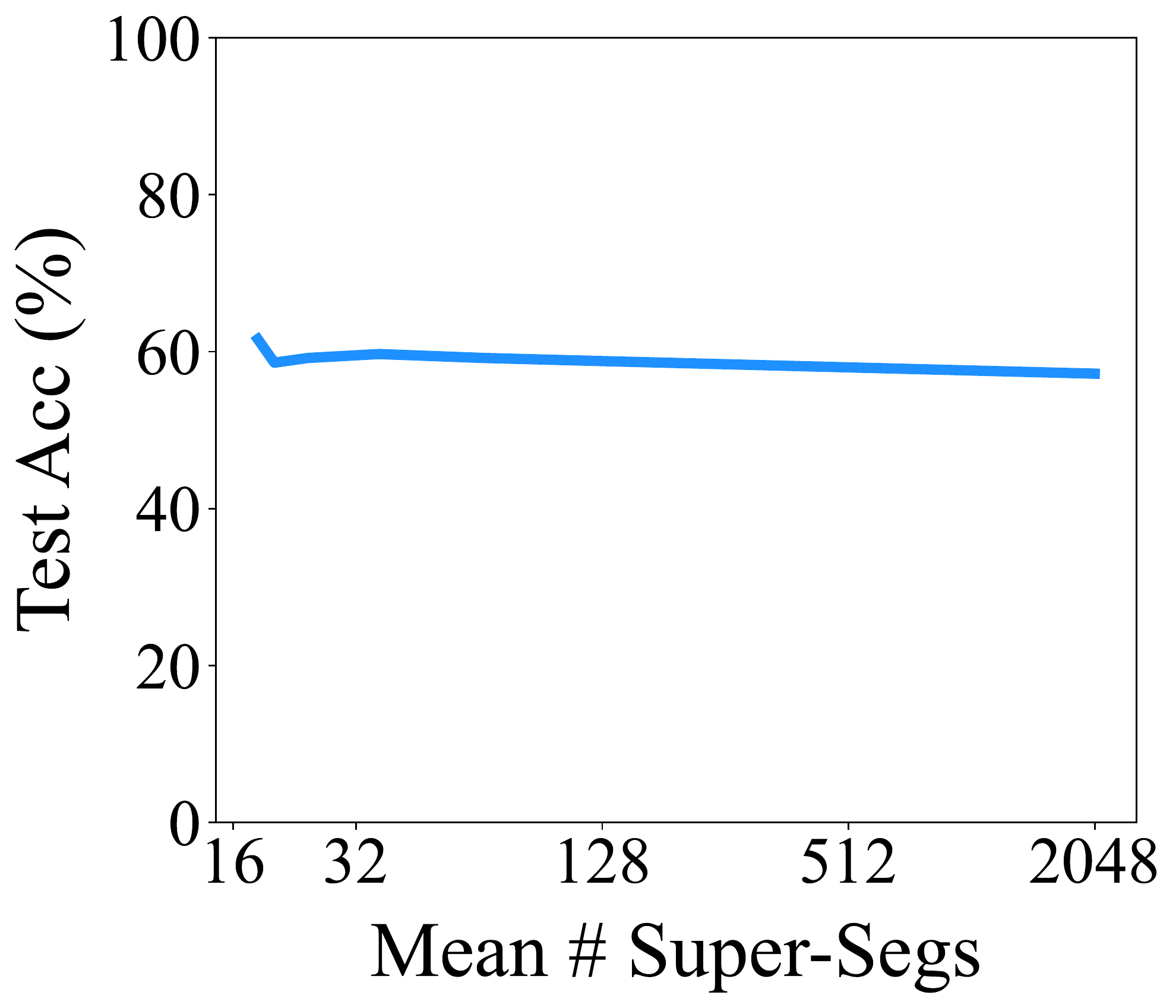}
\vspace{-0.5\baselineskip}
\caption{Results with different granularities of the super-segments. Left shows the average mIoUs, and right shows the target shape classification accuracy. The X-axis is the average number of super-segments in each shape in \emph{log scale}; the higher, the smaller the super-segments are.}
\label{fig:granularity}
\vspace{-0.5\baselineskip}
\end{figure}
Our results in Section~\ref{sec:results}~\refinpaper{} shows that using a set of super-segments as input instead of a point cloud is one of the crucial parts of our framework to achieve meaningful segmentation results through attention, while the accuracy of the target shape classification is not affected much by the representation of shapes. We further analyze the effect of super-segments while varying their granularity.
Although BSP-Net has parameters about the number of planes and the \emph{maximum} number of convexes in training, increasing these numbers does not lead to producing more final convexes in practice.
Thus, to achieve finer granularities, we use K-means clustering implemented in scikit-learn~\cite{scikit-learn} to split each given super-segment into smaller pieces. For each super-segment $\mathbf{s}_i$, we use K-means clustering for the points included in the super-segment (let $\mathcal{P}_i$ denote the points) and set the number of subgroups $K$ in the clustering to be $[ | \mathcal{P}_i | / N ]$, where $N$ is our granularity parameter and $[\cdot]$ is the rounding function. When $N$ is set to be the number of points in the entire point cloud (\num{2048} in our experiments), it is the extreme case that the set of super-segments becomes the input point cloud itself. We test our network (with the PN-Aware setup) while varying the $N$ from 16 to 256 (and 2048, which is the extreme case). Figure~\ref{fig:granularity} illustrates the changes of the average mIoU in the part segmentation (left) and the accuracy of the target shape classification (right). The X-axis of the plots shows the average number of super-segments in each shape in \emph{log scale}, and the Y-axis shows either the average mIoU or the accuracy. Interestingly, the granularity of the super-segments does not make any meaningful difference in the accuracy of the target shape classification but greatly affects the part segmentation mIoUs; more super-segments (smaller super-segments) results in a worse segmentation. This concludes that \emph{pre-merging} the points as much as possible with geometric properties is the key to obtaining meaningful attention maps aligned with semantic parts.

\subsection{Effect of Learning Attention}
\label{sec:effect_learning_attention}
\input{tables/table_acc_across_attention}

To demonstrate whether our neural network learns the attention in a way to improve the discrimination of the target shape, we compare our attention module with two cases: using \emph{uniform} attention maps and using \emph{random} attention maps. As shown in Table~\ref{tbl:all_pairs}, uniform attention maps provide better accuracy in the target shape classification compared with random attention maps, although its accuracy is still lower than the accuracy of our network learning the attention.

\subsection{Out-of-Distribution Test --- More Categories}
\label{sec:ood_test_more}
\input{tables/table_suppl_ood}

In addition to the out-of-distribution test results in Section~\ref{sec:out_of_distribution}~\refinpaper, we provide more results testing our network trained with Chair shapes and utterances to Airplanes and Cars. The mIoUs across the parts are reported in Table~\ref{tbl:suppl_ood}, and qualitative results are in Section~\ref{sec:more_segmentations}. The model trained with the PN-Aware setup is used. Despite the big difference in the shapes, our model still recognizes some semantic parts such as Airplane \emph{body}, \emph{tail}, and \emph{wing} and Car \emph{body} and \emph{wheel}. 

\begin{figure*}[h]
\includegraphics[width=0.99\textwidth]{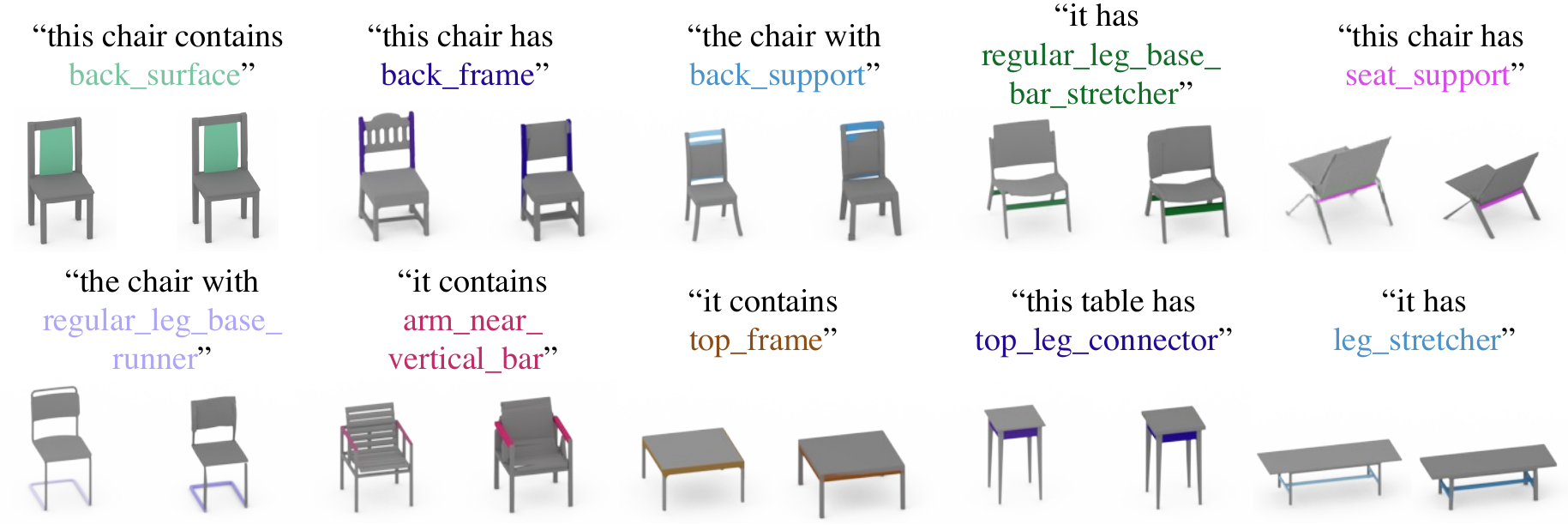}
\vspace{-0.5\baselineskip}
\caption{\textbf{PartNet Results.} For each pair, top is the utterance, left is the ground truth, and right is the predicted segment. The segmented parts of the given utterances are highlighted in color.}
\vspace{-0.5\baselineskip}
\label{fig:partnet}
\end{figure*}

\subsection{Effect of Training Data Size}
\label{sec:effect_training_data_size}
\input{tables/table_sentence_ratio} 
Table~\ref{tbl:data_reduction} illustrates results when only 50\% and 25\% of utterances are used in training (in the PN-Aware case). The part segmentation mIoUs are not changed much even when only 50\% of the utterances are used, even when the target shape classification accuracy is decreased. In an extreme case using only 25\% of the utterances in training, the mIoUs are decreased.
This shows that our network does not necessitate a huge scale of data to obtain meaningful results.

\subsection{Cross-Part mIoUs}
\label{sec:cross_part_mious}
\input{tables/table_mIoU_all_pairs}

We report the cross-part mIoUs for both PN-Agnostic and PN-Aware cases in Table~\ref{tbl:all_pairs}. The diagonals are the same numbers reported in row 2 and 3 in Table~\ref{tbl:results}~\refofpaper{}. For both cases, the mIoUs of the corresponding parts are overwhelmingly higher than the ones of the other parts, indicating that there is almost no overlap across the attention maps of the parts.

\subsection{Low Rank Regularization in AdaCoSeg~\cite{Zhu:2020}}
\label{sec:low_rank_reg}
\input{tables/table_low_rank}

AdaCoSeg~\cite{Zhu:2020} discussed in Section~\ref{sec:related_work}~\refinpaper{} introduces a novel rank-based loss function improving the performance of the co-segmentation task. The loss function called \emph{group consistency loss} (see Section 4.2 in the AdaCoSeg paper) maximizes the similarity of descriptors of the entities (super-segments in our case) included in the same group while differentiating the descriptors of entities assigned to different groups. The loss is computed with entities of \emph{multiple} shapes in a minibatch, and thus it can enforce the consistency of the segmentation across multiple shapes.

We try to adapt this loss function to our network training. 
While this loss does not require labels, it still needs to \emph{define} groups. Hence, to adapt the loss to our network training, we use the PN-Aware setup and consider the sets of super-segments belonging to each predefined part as the groups.
For each super-segment $\mathbf{s_i}$, we take the output of Per-Super Segment Encoder $g(\mathbf{s_i})$, the output of the last layer fed to predict the key $g_k(\mathbf{s_i})$ and value $g_v(\mathbf{s_i})$ vectors. We collate these descriptors for the super-segments assigned to each part (based on the attention outputs); let $M_k$ denote the set of the descriptors (a matrix) for the $k$-th part. The group consistency loss of AdaCoSeg is then defined as follows:

\begin{align}
    \mathcal{L}_{\text{Coseg}}=1+\max_{k}{\mathit{rank}(M_k)}-\min_{k \neq l}{\mathit{rank}([M_k,M_l])},
\end{align}

where $\mathit{rank}$ indicates the second singular value of the matrix and $[\cdot]$ denotes the concatenation of two matrices.

Table~\ref{tbl:reg_loss} shows the results when using the group consistency loss in our network training. We find that the group consistency loss does not help improve the segmentation accuracy in our case. This result implies that the attention module in our network already learns consistent attention maps for each part.

\subsection{Different Utterance Encoder --- ALBERT~\cite{ALBERT}}
\input{tables/table_albert_results}

For the utterance encoding, we also try the other Transformer-based encoder: ALBERT~\cite{ALBERT}, which is a lite version of BERT~\cite{BERT}. We experimented with ALBERT in two ways: using a pretrained model and finetuning it, and without a pretrained model and training from scratch; the pretrained model is obtained from training BookCorpus ~\cite{Zhu:2015} dataset with a masked language model objective. In the PN-Aware setup, the output of the \emph{classification} encoder $f_c(\mathbf{u})$ is obtained from the last hidden state of the [CLS] token, which is further processed through an MLP. The output of the \emph{attention} encoder $f_a(\mathbf{l_k})$ with a part name $\mathbf{l_k}$ is obtained from the word embedding layer of ALBERT and also processed through an MLP. The results are compared in Table~\ref{tbl:albert}.
Interestingly, the finetuned ALBERT does not improve the overall part segmentation mIoUs compared with our case of using a much simpler encoder, LSTM,
while the target shape classification accuracy is higher. When training ALBERT from scratch, the classification accuracy decreases compared to using LSTM.

\subsection{More Fine-grained Parts Segmentation Test}
\label{sec:partnet_results}
\input{tables/table_partnet_results}

In our experiments so far, we used the four part classes (\emph{back}, \emph{seat}, \emph{leg}, \emph{arm}) of Chair in ShapeNet to match the majority of part names used in CiC utterances. To experiment with more parts, we would require a new reference game dataset that contains new part names in descriptions.

Here, we demonstrate experimental results with \emph{synthetic} reference games created using the parts in PartNet~\cite{Mo:2019}. We tested with two categories separately: Chair and Table. Among the parts in \emph{level 2} of the PartNet hierarchy, we randomly sample one of them and synthesize an utterance with template sentences (shown in Figure~\ref{fig:partnet}) indicating the existence or absence of a part. Given the utterance, target and distractor shapes are also randomly sampled based on part existence. Part classes which are present or absent in less than 10\% of shapes are excluded. 
Figure~\ref{fig:partnet} illustrates some qualitative results. Finer-grained parts such as frame, support, and stretcher are accurately discovered. Table~\ref{tbl:partnet_results} also shows the average mIoUs, which are comparable with the \emph{supervised} segmentation result using PointNet, shown in Table 3 of PartNet~\cite{Mo:2019}.
The average mIoUs for both PN-Agnostic and PN-Aware network trained on this dataset are 35.3 and 32.7 respectively, which are comparable with the \emph{supervised} segmentation result using PointNet (38.2) shown in Table 2 of PartNet~\cite{Mo:2019}. (But we also note that the supervised segmentation result in PartNet is the case when using the entire set of level 2 parts without any exclusion.)

\subsection{Implementation Details}
\label{sec:implementation_details}
For Per-Super-Segment Encoder $g$, we used a simplified version PointNet~\cite{pointnet}. Our network takes a set of points included in each super-segment as input and processes the points using 64-dimensional linear layers with BatchNorms and ReLUs. The features of each point are then max-pooled to produce the feature of the super-segment.

In the utterance encoders $f_a(\cdot)$ and $f_c(\cdot)$, the dimensions of the word embedding and the LSTM hidden states are set to 100 and 64, respectively. The word attention method introduced in ShapeGlot~\cite{Achlioptas:2019} is used.
In the cross attention module, a single attention layer is used, which is followed by an MLP and LayerNorm. 

We train our networks for 30 epochs with batch size 64 and use the ADAM ~\cite{Kingma2014} optimizer. The initial learning rate is $10^{-3}$ and decayed by a polynomial scheduler (power=$0.9$). Both regularization losses $\mathcal{L}_{\text{CE}}$ and $\mathcal{L}_{\text{Coseg}}$ are weighted by $10^{-2}$.
When computing cross entropy for the target shape classification and also for the regularization loss $\mathcal{L}_{\text{CE}}$,
we follow ShapeGlot~\cite{Achlioptas:2019} and use the label smoothing technique introduced by Szegedy~\etal~\cite{Szegedy:2016} with the same parameters.

\begin{center}
\textbf{Sections for more qualitative results are\\in the following pages.}
\end{center}

\clearpage
\newpage
\onecolumn
\subsection{More Segmentation Results}
\label{sec:more_segmentations}
In the following, we provide more results of the part segmentation for Chairs, Tables, Lamps, Airplanes, and Cars, as shown in Figure~\ref{fig:teaser}~\refinpaper{}. All the examples in the figure below are \emph{randomly} sampled.
\input{figures/supp_segmentations/supp_segmentations}

\clearpage
\newpage
\onecolumn
\subsection{More Comparisons Results}
\label{sec:more_comparisons}
We also provide more results of the comparison with the other methods below, as shown in Figure~\ref{fig:results}~\refinpaper{}. All the examples in the figure below are \emph{randomly} sampled.

\input{figures/supp_comparisons/supp_comparisons}

\clearpage
\newpage
\twocolumn

%% file: tables/table_acc_across_attention.tex
\begin{table}[h!]
\centering
\vspace{-0.5\baselineskip}
\caption{Comparison with the cases of using uniform and random attention maps.}
\footnotesize{
\setlength{\tabcolsep}{0.2em}
\renewcommand{\arraystretch}{1.0}
\begin{tabularx}{\linewidth}{Y|Y}
  \toprule

Method & Classification Accuracy (\%) \\
\midrule
Random $\{ \mathbf{w_i} \}$ & 58.4\\ 
Uniform $\{ \mathbf{w_i} \}$ & 59.3\\ 
Ours (Learning $\{ \mathbf{w_i} \}$) & \textbf{61.5}\\ 
  \bottomrule
\end{tabularx}
}
\vspace{-0.5\baselineskip}
\label{tbl:acc_comparison}
\end{table}

%% file: tables/table_suppl_ood.tex
\begin{table}[h!]
\centering
\vspace{-0.5\baselineskip}
\caption{Quantitative results of the out-of-distribution test with Airplanes and Cars. The highest mIoU for each part of the target class is marked in \textbf{bold}.}
\footnotesize{
\setlength{\tabcolsep}{0.2em}
\renewcommand{\arraystretch}{1.0}
\begin{tabularx}{\linewidth}{Y|>{\centering}m{1.6cm}|Y|Y|Y|Y}
  \toprule
   \multicolumn{2}{c|}{\multirow{2}{*}{\makecell{Other\\Classes}}} & \multicolumn{4}{c}{Chair (w/ PN-Aware)} \\
   \cline{3-6}
   \multicolumn{2}{c|}{} & Back & Seat & Leg & Arm \\
  
   \midrule
   \multirow{4}{*}{Airplane} & Body & 17.5 & 26.1 & \textbf{30.2} & 0.2 \\
   & Wing & 3.1 & \textbf{47.5} & 3.5 & 6.3 \\
   & Tail & \textbf{46.5} & 0.8 & 1.0 & 0.2 \\
   & Engine & 5.4 & \textbf{11.6} & 6.1 & 7.5 \\
  
   \midrule
   \multirow{4}{*}{Car} & Roof & 6.2 & 2.8 & 0.6 & \textbf{7.5} \\
   & Hood & 0.1 & \textbf{17.5} & 1.5 & 0.6 \\
   & Wheel & 9.9 & 12.5 & \textbf{21.3} & 1.6 \\ 
   & Body & \textbf{45.3} & 29.5 & 2.4 & 10.5 \\
  
  \bottomrule
\end{tabularx}
}
\vspace{-0.5\baselineskip}
\label{tbl:suppl_ood}
\end{table}

%% file: tables/table_sentence_ratio.tex
\begin{table}[h!]
\centering
\vspace{-0.5\baselineskip}
\caption{Results when training with a subset of the training data. \textbf{Bold} indicates the best result for each column.}
\footnotesize{
\setlength{\tabcolsep}{0.2em}
\renewcommand{\arraystretch}{1.0}
\begin{tabularx}{\linewidth}{>{\centering}m{1.4cm}|Y|Y|Y|Y|Y|Y}
  \toprule
  \multirow{2}{*}{\makecell{Utterance\\Rate}} &
  \multicolumn{5}{c|}{Segmentation mIoU(\%)} &
  \multirow{2}{*}{\makecell{Classif.\\Acc.(\%)}}\\
  \cline{2-6}
  & Back & Seat & Leg & Arm & Avg. & \\
  \midrule
  $100\%$ & \textbf{84.9} & \textbf{83.6} & \textbf{78.9} & 70.4 & \textbf{79.4} & \textbf{61.5} \\
  $50\%$ & 80.9 & 79.0 & 77.1 & \textbf{70.9} & 77.0 & 56.0 \\
  $25\%$ & 56.5 & 37.5 & 76.1 & 66.1 & 59.1 & 53.8 \\

  \bottomrule
\end{tabularx}
}
\vspace{-0.5\baselineskip}
\label{tbl:data_reduction}
\end{table}

%% file: tables/table_mIoU_all_pairs.tex
\begin{table}[h!]
\centering
\vspace{-0.5\baselineskip}
\caption{Part segmentation mIoUs across parts. The highest mIoU for each ground truth part is marked in \textbf{bold}.}
\footnotesize{
\setlength{\tabcolsep}{0.2em}
\renewcommand{\arraystretch}{1.0}
\begin{tabularx}{\linewidth}{Y|Y|Y|Y|Y}
  \toprule
  \multirow{2}{*}{\makecell{Ground\\Truth}} & \multicolumn{4}{c}{Prediction} \\
  \cline{2-5}
  & Back & Seat & Leg & Arm \\
  
  \midrule
  \multicolumn{5}{c}{PN-Agnostic (Sec.~\ref{sec:pn_free})}\\
  \midrule
  Back & \textbf{82.2} & 4.2 & 1.6 & 3.5 \\
  Seat & 0.8 & \textbf{78.8} & 1.5 & 5.2 \\
  Leg & 0.5 & 4.2 & \textbf{75.5} & 3.1 \\
  Arm & 0.2 & 0.7 & 0.8 & \textbf{40.6} \\
  
  \midrule
  \multicolumn{5}{c}{PN-Aware (Sec.~\ref{sec:pn_aware})}\\
  \midrule
  Back & \textbf{84.9} & 2.5 & 1.5 & 1.7 \\
  Seat & 1.8 & \textbf{83.6} & 2.6 & 1.4 \\
  Leg & 1.1 & 2.4 & \textbf{78.9} & 0.7 \\
  Arm & 0.4 & 0.6 & 1.3 & \textbf{70.4} \\
  
  \bottomrule
\end{tabularx}
}
\vspace{-0.5\baselineskip}
\label{tbl:all_pairs}
\end{table}

%% file: tables/table_low_rank.tex
\begin{table}[h!]
\centering
\vspace{-0.5\baselineskip}
\caption{Results with the group consistency loss introduced by AdaCoSeg~\cite{Zhu:2020}. \textbf{Bold} indicates the best result for each column.}
\newcolumntype{Y}{>{\centering\arraybackslash}X}
\footnotesize{
\setlength{\tabcolsep}{0.2em}
\renewcommand{\arraystretch}{1.0}
\begin{tabularx}{\linewidth}{>{\centering}m{1.8cm}|Y|Y|Y|Y|Y|Y}
  \toprule
  \multirow{2}{*}{Regularization} &
  \multicolumn{5}{c|}{Segmentation mIoU(\%)} &
  \multirow{2}{*}{\makecell{Classif.\\Acc.(\%)}}\\
  \cline{2-6}
  & Back & Seat & Leg & Arm & Avg. & \\

  \midrule
  $\mathcal{L}_{\text{CE}}$ (Ours) & \textbf{84.9} & \textbf{83.6} & 78.9 & 70.4 & \textbf{79.4} & \textbf{61.5} \\ 
  $\mathcal{L}_{\text{CE}}$ + $\mathcal{L}_{\text{Coseg}}$ & 83.4 & 82.2 & \textbf{79.2} & 72.0 & 79.2 & 60.2 \\
  $\mathcal{L}_{\text{Coseg}}$ & 79.2 & 80.1 & 78.1 & \textbf{72.5} & 77.5 & 60.6 \\ 

  \bottomrule
\end{tabularx}
}
\vspace{-0.5\baselineskip}
\label{tbl:reg_loss}
\end{table}

%% file: tables/table_albert_results.tex
\begin{table}[h!]
\centering
\vspace{-0.5\baselineskip}
\caption{Results with ALBERT~\cite{ALBERT} as utterance encoders. \textbf{Bold} indicates the best result for each column.}

\newcolumntype{Y}{>{\centering\arraybackslash}X}
\footnotesize{
\setlength{\tabcolsep}{0.1em}
\renewcommand{\arraystretch}{1.0}
\begin{tabularx}{\linewidth}{>{\centering}m{2.2cm}|Y|Y|Y|Y|Y|Y}
  \toprule
  \multirow{2}{*}{\makecell{Utterance\\Encoder}} &
  \multicolumn{5}{c|}{Segmentation mIoU(\%)} &
  \multirow{2}{*}{\makecell{Classif.\\Acc.(\%)}}\\
  \cline{2-6}
  & Back & Seat & Leg & Arm & Avg. & \\
  \midrule 
  ALBERT (w/ FT) & 83.1 & 81.5 & \textbf{79.7} & 61.1 & 76.4 & \textbf{62.9} \\ 
  ALBERT (w/o FT)  & 80.9 & 80.8 & 78.7 & \textbf{72.6} & 78.2 & 57.8 \\
  LSTM (Ours) & \textbf{84.9} & \textbf{83.6} & 78.9 & 70.4 & \textbf{79.4} & 61.5 \\ 

  \bottomrule
\end{tabularx}
}
\vspace{-0.5\baselineskip}
\label{tbl:albert}
\end{table}

%% file: tables/table_partnet_results.tex
\begin{table}[h!]
\vspace{-0.5\baselineskip}
\centering
\caption{The average mIoUs with \emph{level 2} parts of Chair and Table in PartNet~\cite{Mo:2019}.
Due to the lack of utterances in the CiC dataset including words for the finer-grained parts, synthetic utterances are used.
The results in the third column are the case shown in PartNet~\cite{Mo:2019} when the network is \emph{fully supervised} with the ground truth segments. Compared with that, our network learning only from referential language shows comparable results.
For each category, the highest mIoU is marked in \textbf{bold}.}
{
\footnotesize
\setlength{\tabcolsep}{0.2em}
\renewcommand{\arraystretch}{1.0}
\begin{tabularx}{\linewidth}{Y|YYY}
\toprule
Category & PN-Agnostic & PN-Aware & \makecell{Supervised~\cite{Mo:2019}} \\
\midrule
Chair & 35.3 & 32.7 & \textbf{38.2} \\
Table & \textbf{44.0} & 33.7 & 34.3 \\
\bottomrule
\end{tabularx}
\centering 
}
\vspace{-0.5\baselineskip}
\label{tbl:partnet_results}
\end{table}

%% file: figures/supp_segmentations/supp_segmentations.tex
\newcommand{\AllSegmentationImages}{

\input{figures/supp_segmentations/image_list.tex}}
\graphicspath{{figures/supp_segmentations/}}

\makeatletter
\def\Image#1{%
  \multicolumn{\LT@cols}{l}{\includegraphics[width=0.77\textwidth]{#1}}\\
}
\makeatother

\setlength{\tabcolsep}{0em}
\def\arraystretch{0.0}
\newcolumntype{Z}{>{\centering\arraybackslash}m{0.11\textwidth}}
{\scriptsize
\begin{longtable}{Z|ZZZZ|Z|Z}
\makecell{Input\\(Super-Seg.)} &
Back &
Seat &
Leg &
Arm &
\makecell{\textbf{Output}\\Segments} &
GT\\
  \midrule
  \endhead

  \bottomrule
  \endfoot

\input{figures/supp_segmentations/image_list.tex}

\end{longtable}
}

%% file: figures/supp_segmentations/image_list.tex
\Image{figures/\suppSegDir/chair_00.png}
\Image{figures/\suppSegDir/chair_01.png}
\Image{figures/\suppSegDir/chair_02.png}
\Image{figures/\suppSegDir/chair_03.png}
\Image{figures/\suppSegDir/chair_04.png}
\Image{figures/\suppSegDir/chair_05.png}
\Image{figures/\suppSegDir/chair_06.png}
\Image{figures/\suppSegDir/chair_07.png}
\Image{figures/\suppSegDir/chair_08.png}
\Image{figures/\suppSegDir/chair_09.png}
\Image{figures/\suppSegDir/chair_10.png}

\Image{figures/\suppSegDir/chair_25.png}
\Image{figures/\suppSegDir/chair_26.png}
\Image{figures/\suppSegDir/chair_27.png}
\Image{figures/\suppSegDir/chair_28.png}
\Image{figures/\suppSegDir/chair_29.png}
\Image{figures/\suppSegDir/chair_30.png}
\Image{figures/\suppSegDir/chair_31.png}
\Image{figures/\suppSegDir/chair_32.png}
\Image{figures/\suppSegDir/chair_33.png}
\Image{figures/\suppSegDir/chair_34.png}
\Image{figures/\suppSegDir/chair_35.png}
\Image{figures/\suppSegDir/chair_36.png}
\Image{figures/\suppSegDir/chair_37.png}
\Image{figures/\suppSegDir/chair_38.png}

\Image{figures/\suppSegDir/table_00.png}
\Image{figures/\suppSegDir/table_01.png}
\Image{figures/\suppSegDir/table_02.png}
\Image{figures/\suppSegDir/table_03.png}
\Image{figures/\suppSegDir/table_04.png}
\Image{figures/\suppSegDir/table_05.png}
\Image{figures/\suppSegDir/table_06.png}
\Image{figures/\suppSegDir/table_07.png}
\Image{figures/\suppSegDir/table_08.png}
\Image{figures/\suppSegDir/table_09.png}
\Image{figures/\suppSegDir/table_10.png}
\Image{figures/\suppSegDir/table_11.png}
\Image{figures/\suppSegDir/table_12.png}
\Image{figures/\suppSegDir/table_13.png}
\Image{figures/\suppSegDir/table_14.png}
\Image{figures/\suppSegDir/table_15.png}
\Image{figures/\suppSegDir/table_16.png}
\Image{figures/\suppSegDir/table_17.png}
\Image{figures/\suppSegDir/table_18.png}
\Image{figures/\suppSegDir/table_19.png}
\Image{figures/\suppSegDir/table_20.png}
\Image{figures/\suppSegDir/table_21.png}
\Image{figures/\suppSegDir/table_22.png}
\Image{figures/\suppSegDir/table_23.png}
\Image{figures/\suppSegDir/table_24.png}
\Image{figures/\suppSegDir/table_25.png}
\Image{figures/\suppSegDir/table_26.png}
\Image{figures/\suppSegDir/table_27.png}
\Image{figures/\suppSegDir/table_28.png}
\Image{figures/\suppSegDir/table_29.png}
\Image{figures/\suppSegDir/table_30.png}
\Image{figures/\suppSegDir/table_31.png}
\Image{figures/\suppSegDir/table_32.png}
\Image{figures/\suppSegDir/table_33.png}
\Image{figures/\suppSegDir/table_34.png}
\Image{figures/\suppSegDir/table_35.png}
\Image{figures/\suppSegDir/table_36.png}
\Image{figures/\suppSegDir/table_37.png}
\Image{figures/\suppSegDir/table_38.png}

\Image{figures/\suppSegDir/lamp_02.png}
\Image{figures/\suppSegDir/lamp_03.png}
\Image{figures/\suppSegDir/lamp_04.png}
\Image{figures/\suppSegDir/lamp_05.png}
\Image{figures/\suppSegDir/lamp_06.png}
\Image{figures/\suppSegDir/lamp_07.png}
\Image{figures/\suppSegDir/lamp_08.png}
\Image{figures/\suppSegDir/lamp_09.png}
\Image{figures/\suppSegDir/lamp_10.png}
\Image{figures/\suppSegDir/lamp_11.png}
\Image{figures/\suppSegDir/lamp_12.png}
\Image{figures/\suppSegDir/lamp_13.png}
\Image{figures/\suppSegDir/lamp_14.png}
\Image{figures/\suppSegDir/lamp_15.png}
\Image{figures/\suppSegDir/lamp_16.png}
\Image{figures/\suppSegDir/lamp_17.png}
\Image{figures/\suppSegDir/lamp_18.png}
\Image{figures/\suppSegDir/lamp_19.png}
\Image{figures/\suppSegDir/lamp_20.png}
\Image{figures/\suppSegDir/lamp_21.png}
\Image{figures/\suppSegDir/lamp_22.png}
\Image{figures/\suppSegDir/lamp_23.png}
\Image{figures/\suppSegDir/lamp_24.png}
\Image{figures/\suppSegDir/lamp_25.png}
\Image{figures/\suppSegDir/lamp_26.png}
\Image{figures/\suppSegDir/lamp_27.png}
\Image{figures/\suppSegDir/lamp_28.png}
\Image{figures/\suppSegDir/lamp_29.png}
\Image{figures/\suppSegDir/lamp_30.png}
\Image{figures/\suppSegDir/lamp_31.png}
\Image{figures/\suppSegDir/lamp_32.png}
\Image{figures/\suppSegDir/lamp_33.png}
\Image{figures/\suppSegDir/lamp_34.png}
\Image{figures/\suppSegDir/lamp_35.png}
\Image{figures/\suppSegDir/lamp_36.png}

\Image{figures/\suppSegDir/airplane_00.png}
\Image{figures/\suppSegDir/airplane_01.png}
\Image{figures/\suppSegDir/airplane_02.png}
\Image{figures/\suppSegDir/airplane_03.png}
\Image{figures/\suppSegDir/airplane_04.png}
\Image{figures/\suppSegDir/airplane_05.png}
\Image{figures/\suppSegDir/airplane_06.png}
\Image{figures/\suppSegDir/airplane_07.png}
\Image{figures/\suppSegDir/airplane_08.png}
\Image{figures/\suppSegDir/airplane_09.png}
\Image{figures/\suppSegDir/airplane_10.png}
\Image{figures/\suppSegDir/airplane_11.png}
\Image{figures/\suppSegDir/airplane_12.png}
\Image{figures/\suppSegDir/airplane_13.png}
\Image{figures/\suppSegDir/airplane_14.png}
\Image{figures/\suppSegDir/airplane_15.png}
\Image{figures/\suppSegDir/airplane_16.png}
\Image{figures/\suppSegDir/airplane_17.png}
\Image{figures/\suppSegDir/airplane_18.png}
\Image{figures/\suppSegDir/airplane_19.png}
\Image{figures/\suppSegDir/airplane_20.png}
\Image{figures/\suppSegDir/airplane_21.png}
\Image{figures/\suppSegDir/airplane_22.png}
\Image{figures/\suppSegDir/airplane_23.png}
\Image{figures/\suppSegDir/airplane_24.png}
\Image{figures/\suppSegDir/airplane_25.png}
\Image{figures/\suppSegDir/airplane_26.png}
\Image{figures/\suppSegDir/airplane_27.png}
\Image{figures/\suppSegDir/airplane_28.png}
\Image{figures/\suppSegDir/airplane_29.png}
\Image{figures/\suppSegDir/airplane_30.png}
\Image{figures/\suppSegDir/airplane_31.png}
\Image{figures/\suppSegDir/airplane_32.png}
\Image{figures/\suppSegDir/airplane_33.png}
\Image{figures/\suppSegDir/airplane_34.png}
\Image{figures/\suppSegDir/airplane_35.png}
\Image{figures/\suppSegDir/airplane_36.png}
\Image{figures/\suppSegDir/airplane_37.png}

\Image{figures/\suppSegDir/car_00.png}
\Image{figures/\suppSegDir/car_01.png}
\Image{figures/\suppSegDir/car_02.png}
\Image{figures/\suppSegDir/car_03.png}
\Image{figures/\suppSegDir/car_04.png}
\Image{figures/\suppSegDir/car_05.png}
\Image{figures/\suppSegDir/car_06.png}
\Image{figures/\suppSegDir/car_07.png}
\Image{figures/\suppSegDir/car_08.png}
\Image{figures/\suppSegDir/car_09.png}
\Image{figures/\suppSegDir/car_10.png}
\Image{figures/\suppSegDir/car_11.png}
\Image{figures/\suppSegDir/car_12.png}
\Image{figures/\suppSegDir/car_13.png}
\Image{figures/\suppSegDir/car_14.png}
\Image{figures/\suppSegDir/car_15.png}
\Image{figures/\suppSegDir/car_16.png}
\Image{figures/\suppSegDir/car_17.png}
\Image{figures/\suppSegDir/car_18.png}
\Image{figures/\suppSegDir/car_19.png}
\Image{figures/\suppSegDir/car_20.png}
\Image{figures/\suppSegDir/car_21.png}
\Image{figures/\suppSegDir/car_22.png}
\Image{figures/\suppSegDir/car_23.png}
\Image{figures/\suppSegDir/car_24.png}
\Image{figures/\suppSegDir/car_25.png}
\Image{figures/\suppSegDir/car_26.png}
\Image{figures/\suppSegDir/car_28.png}

%% file: figures/supp_comparisons/supp_comparisons.tex
\newcommand{\AllComparisonImages}{\input{figures/supp_comparisons/image_list.tex}}
\graphicspath{{figures/supp_comparisons/}}

\makeatletter
\def\Image#1{%
  \multicolumn{\LT@cols}{l}{\includegraphics[width=\textwidth]{#1}}\\
}
\makeatother

\setlength{\tabcolsep}{0em}
\def\arraystretch{0.0}
\newcolumntype{Z}{>{\centering\arraybackslash}m{0.0909\textwidth}}
{\scriptsize
\begin{longtable}{Z|ZZ|ZZZZZZZZ}
GT &
\makecell{PN-Agnostic\\(Ours)} &
\makecell{PN-Aware\\(Ours)} &
Points &
P $\rightarrow$ Sp.-Seg. &
\makecell{w/o\\Unit Norm}&
$\sigma(\mathbf{X}) \rightarrow i$ &
$\sigma(\mathbf{X}) \rightarrow k$ &
\makecell{$\sigma(\mathbf{X})$\\$\rightarrow i \rightarrow k$} &
w/ Global Feat. &
w/o $\mathcal{L}_{\text{CE}}$ \\
  \midrule
  \endhead

  \bottomrule
  \endfoot

  \input{figures/supp_comparisons/image_list.tex}

\end{longtable}
}

%% file: figures/supp_comparisons/image_list.tex
\Image{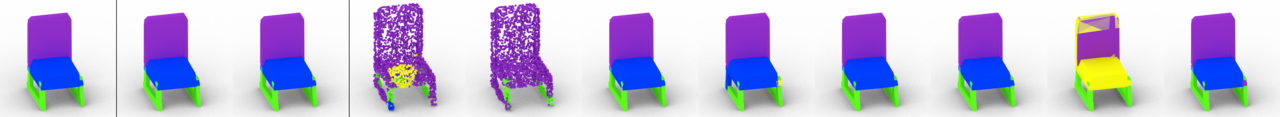}
\Image{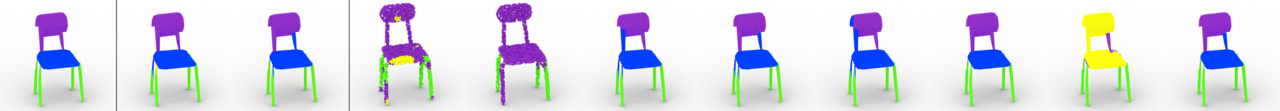}
\Image{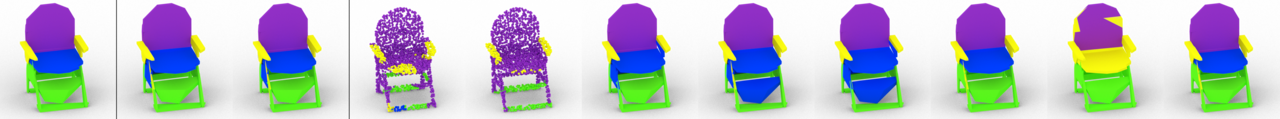}
\Image{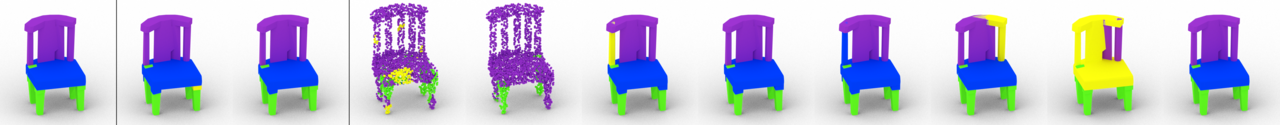}
\Image{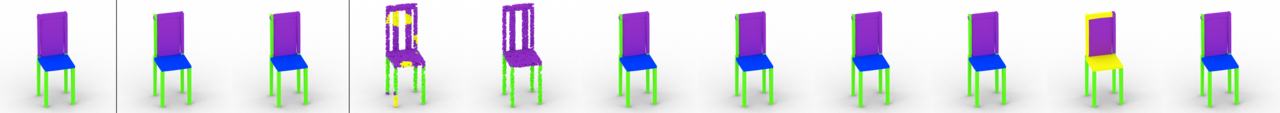}
\Image{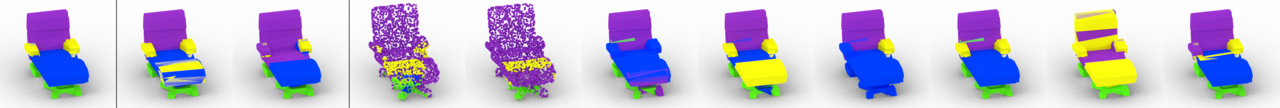}
\Image{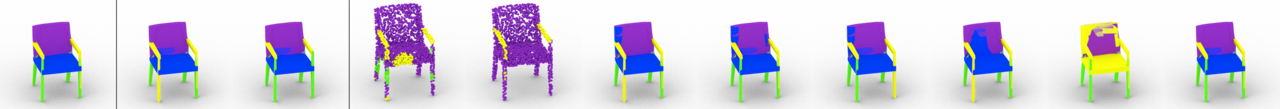}
\Image{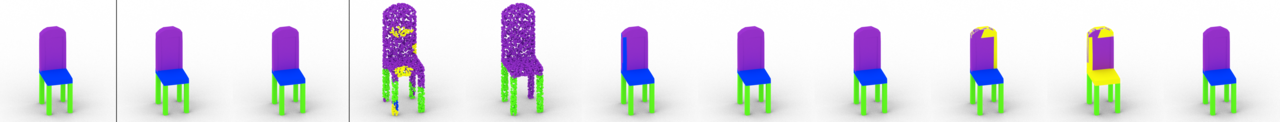}
\Image{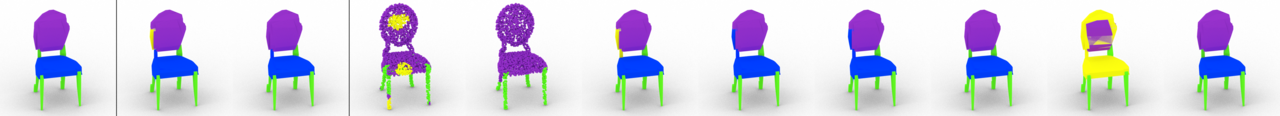}
\Image{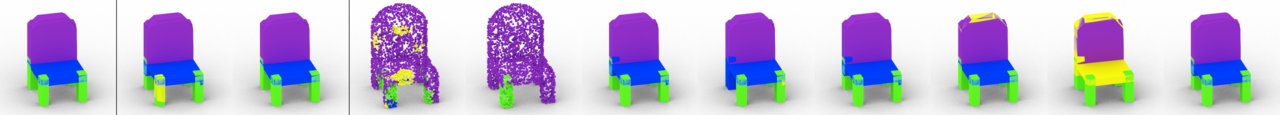}
\Image{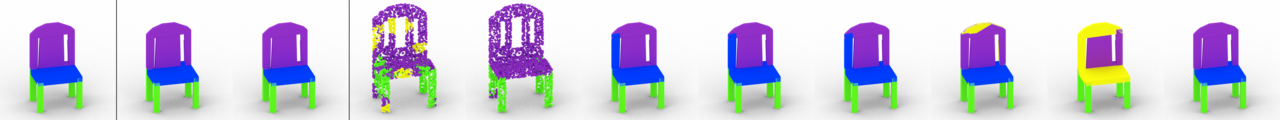}
\Image{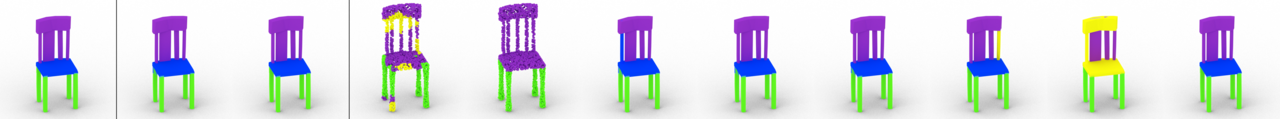}
\Image{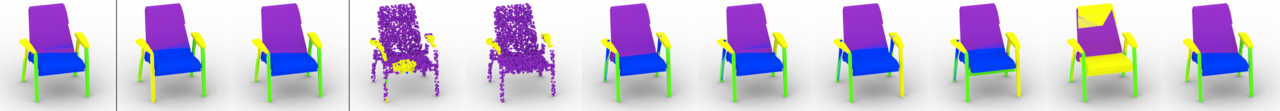}
\Image{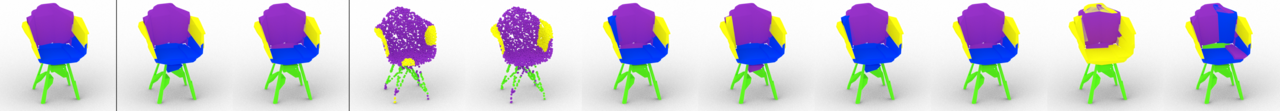}
\Image{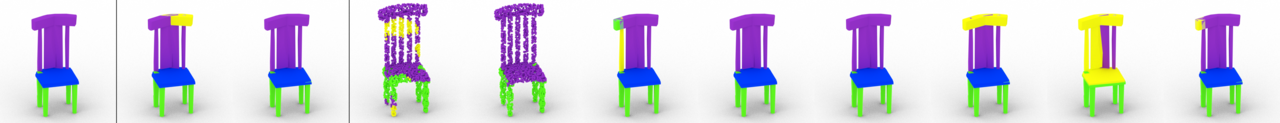}
\Image{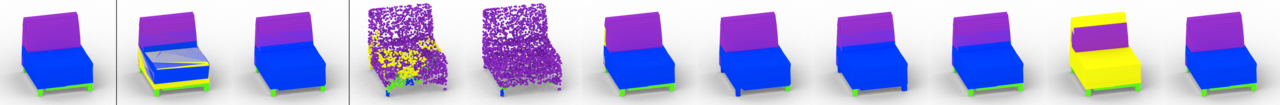}
\Image{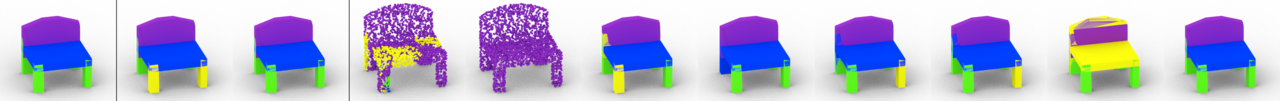}
\Image{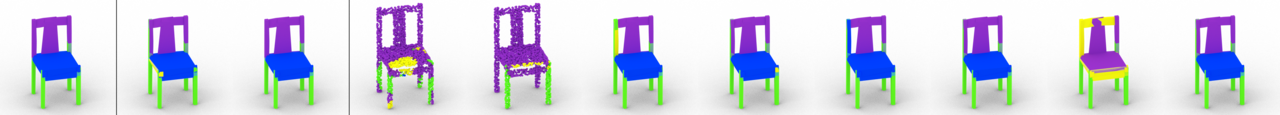}
\Image{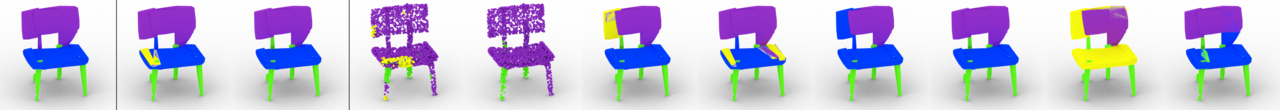}
\Image{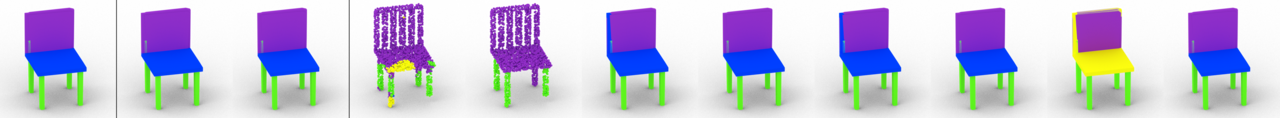}
\Image{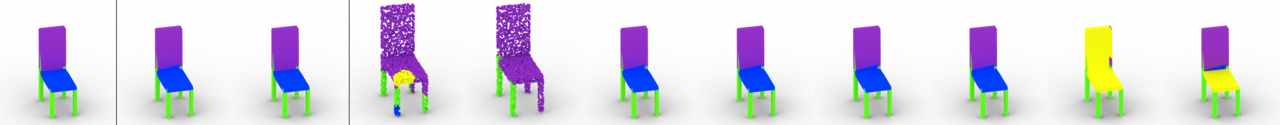}
\Image{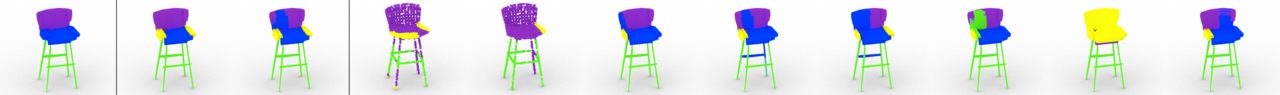}
\Image{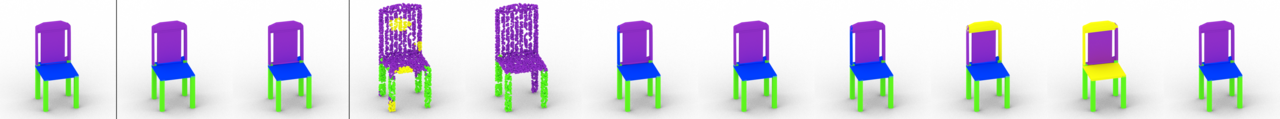}
\Image{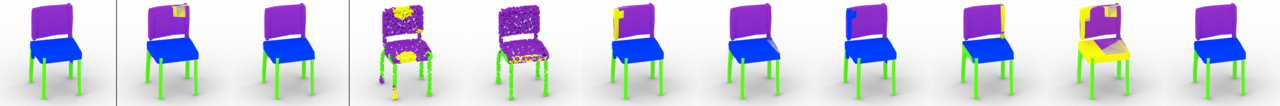}
\Image{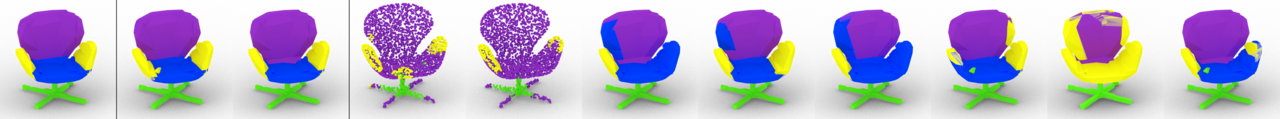}
\Image{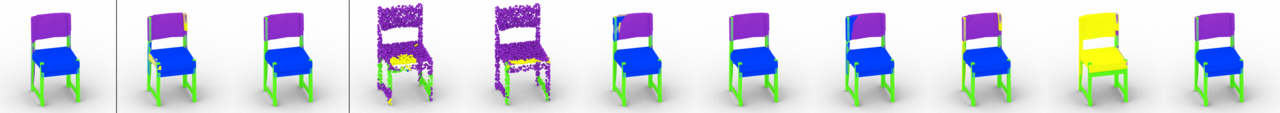}
\Image{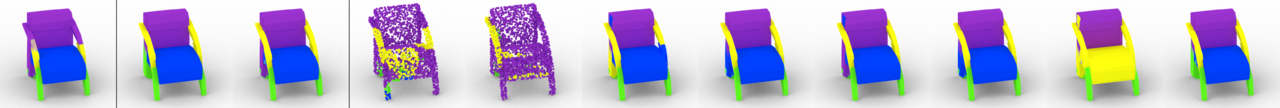}
\Image{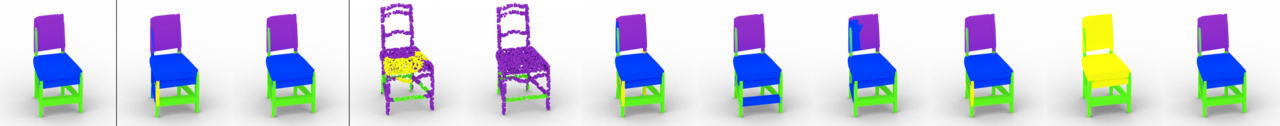}
\Image{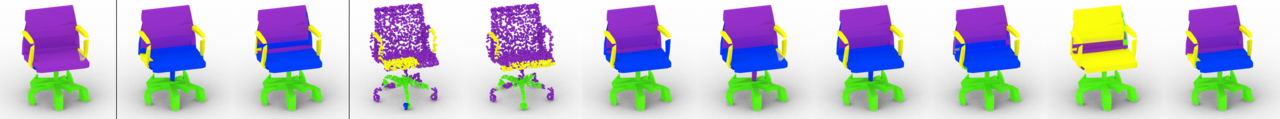}
\Image{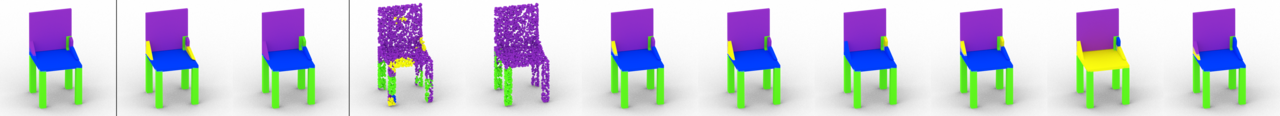}
\Image{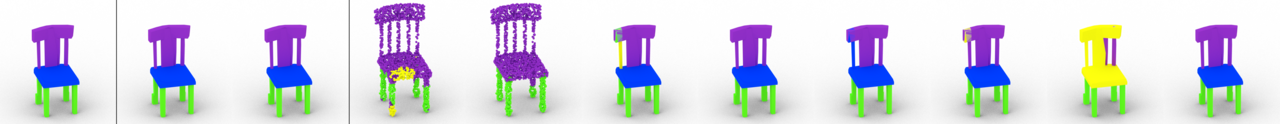}
\Image{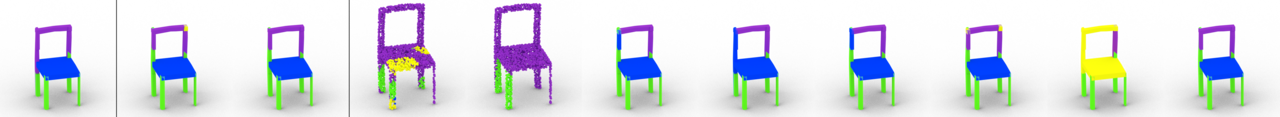}
\Image{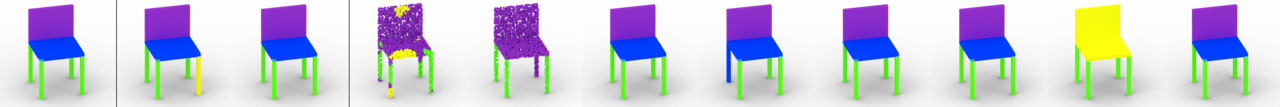}
\Image{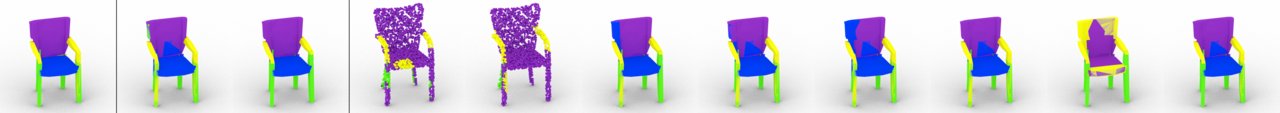}
\Image{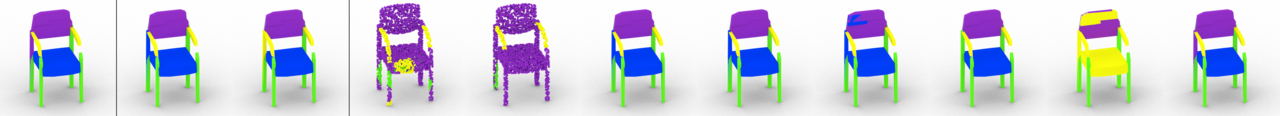}
\Image{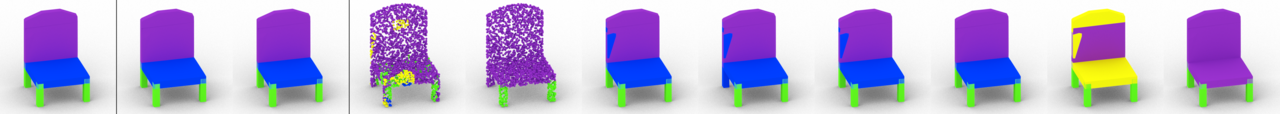}
\Image{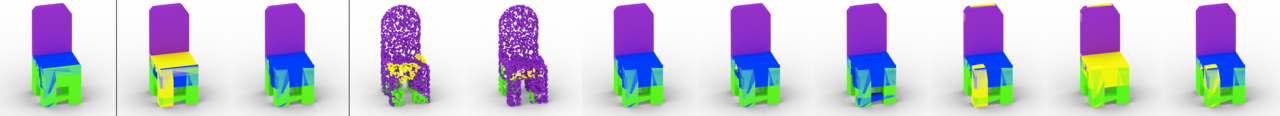}
\Image{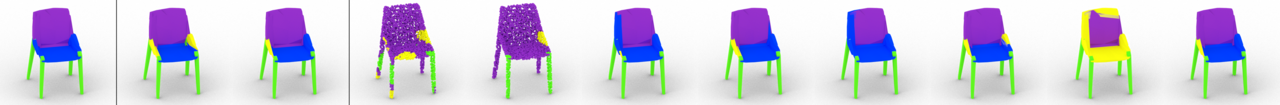}
\Image{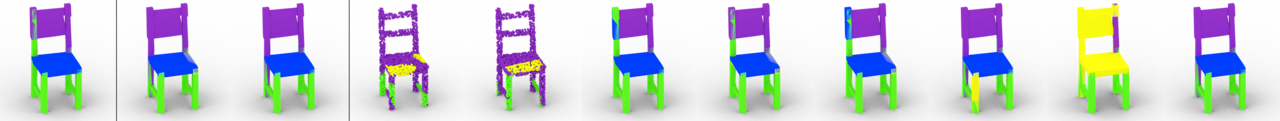}